\documentclass{bmcart}

\usepackage{amsthm,amsmath}
\usepackage[utf8]{inputenc} 

\usepackage{lineno}
\usepackage{graphicx}
\usepackage{subfigure}
\usepackage{caption}

\usepackage{todonotes}
\usepackage{hyperref}
\usepackage{url}
\usepackage{booktabs}
\usepackage{array}
\newcolumntype{H}{>{\setbox0=\hbox\bgroup}c<{\egroup}@{}}

\usepackage{multirow}

\usepackage{amsmath}
\usepackage{xspace}

\startlocaldefs
\newcommand{\novelh}{\textsc{Novel H1N1}\xspace}
\newcommand{\novele}{\textsc{Novel Ebola}\xspace}
\newcommand{\zhouh}{\textsc{Zhou's H1N1}\xspace}
\newcommand{\zhoue}{\textsc{Zhou's Ebola}\xspace}
\newcommand{\barman}{\textsc{Barman}\xspace}
\newcommand{\slim}{\textsc{Denovo\_slim}\xspace}
\newcommand{\generalized}{\textsc{Generalized}\xspace}
\newcommand{\docvec}{\textsc{Doc2vec}\xspace}
\newcommand{\denovo}{\textsc{Denovo}\xspace}
\newcommand{\deepviral}{\textsc{DeepViral}\xspace}
\newcommand{\motif}{\textsc{MotifTransformer}\xspace}
\newcommand{\mtt}{\textsc{MTT}\xspace}
\newcommand{\stt}{\textsc{STT}\xspace}

\newcommand{\sara}{\textsc{SARS-CoV}\xspace}
\newcommand{\sarb}{\textsc{SARS-CoV-2}\xspace}
\newcommand{\naive}{\textsc{Naive baseline}\xspace}

\newcommand{\mpara}[1]{\medskip\noindent{\bf #1}}
\newtheoremstyle{named}{}{}{\itshape}{}{\bfseries}{.}{.5em}{\thmnote{#3}}
\theoremstyle{named}

\endlocaldefs

\begin{document}

\begin{frontmatter}

\begin{fmbox}
\dochead{Methodology}


\title{A multitask transfer learning framework for the prediction of virus-human protein-protein interactions }


\author[
  addressref={aff1},                   
  corref={aff1},                       
  email={dong@l3s.de}   
]{\inits{T. Ngan}\fnm{Thi Ngan} \snm{Dong}}
\author[
  addressref={aff2,aff3},
  email={Graham.Brogden@tiho-hannover.de}
]{\inits{Graham}\fnm{Graham} \snm{Brogden}}
\author[
  addressref={aff2,aff3,aff4,aff5},
  email={Gisa.Gerold@tiho-hannover.de}
]{\inits{Gisa}\fnm{Gisa} \snm{Gerold}}
\author[
  addressref={aff1},
  email={khosla@l3s.de}
]{\inits{Megha}\fnm{Megha} \snm{Khosla}}


\address[id=aff1]{
  \orgdiv{L3S Research Center},             
  \orgname{Leibniz University Hannover},          
  \city{Hannover},                              
  \cny{Germany}                                    
}

\address[id=aff2]{
  \orgdiv{Institute for Biochemistry},             
  \orgname{University of Veterinary Medicine},          
  \city{Hannover},                              
  \cny{Germany}                                    
}
\address[id=aff3]{
  \orgdiv{Institute of Experimental Virology},             
  \orgname{TWINCORE, Center for Experimental and Clinical Infection Research Hannover},          
  \city{Hannover},                              
  \cny{Germany}                                    
}
\address[id=aff4]{
  \orgdiv{Department of Clinical Microbiology},             
  \orgname{Umeå University},          
  \city{Umeå},                              
  \cny{Sweden}                                    
}
\address[id=aff5]{
  \orgdiv{Wallenberg Centre for Molecular Medicine (WCMM)},             
  \orgname{Umeå University},          
  \city{Umeå},                              
  \cny{Sweden}                                    
}



\end{fmbox}


\begin{abstractbox}

\begin{abstract} 

\parttitle{Background} 
Viral infections are causing significant morbidity and mortality worldwide.  Understanding the interaction patterns between a particular virus and human proteins plays a crucial role in unveiling the underlying mechanism of viral infection and pathogenesis. This could further help in prevention and treatment of virus-related diseases. However, the task of predicting protein-protein interactions between a new virus and human cells is extremely challenging due to scarce data on virus-human interactions and fast mutation rates of most viruses.\\
\parttitle{Results}
We developed a \emph{multitask transfer learning} approach that exploits the information of around 24 million protein sequences and the interaction patterns from the human interactome to counter the problem of small training datasets. Instead of using hand-crafted protein features, we utilize statistically rich protein representations learned by a deep language modeling approach from a massive source of protein sequences. Additionally, we employ an additional objective which aims to maximize the probability of observing human protein-protein interactions. This additional task objective acts as a regularizer and also allows to incorporate domain knowledge to inform the virus-human protein-protein interaction prediction model.\\
\parttitle{Conclusions}
Our approach achieved competitive results on 13 benchmark datasets and the case study for the \sarb virus receptor. Experimental results show that our proposed model works effectively for both virus-human and bacteria-human protein-protein interaction prediction tasks. We share our code for reproducibility and future research at~\href{https://git.l3s.uni-hannover.de/dong/multitask-transfer}{\url{https://git.l3s.uni-hannover.de/dong/multitask-transfer}}.

\end{abstract}


\begin{keyword}
\kwd{protein-protein interaction}
\kwd{human PPI}
\kwd{virus-human PPI}
\kwd{multitask}
\kwd{transfer learning}
\kwd{protein embedding}
\end{keyword}

\end{abstractbox}

\end{frontmatter}

\section{Introduction}
\label{secIntroduction}

Virus infections cause an enormous and ever increasing burden on healthcare systems worldwide. The ongoing COVID-19 pandemic caused by the zoonotic virus, SARS-CoV-2, has resulted in enormous socio-economic losses~\cite{socialcostsCOVID-19}. Viruses infect all life forms and require host cells to complete their replication cycle by utilizing the host cell machinery.  Virus infection involves several types of protein-protein interactions (PPIs) between the virus and its host. These interactions include the initial attachment of virus coat or envelope proteins to host membrane receptors, hijacking of the host translation and intracellular transport machineries resulting in replication, assembly and subsequent release of virus particles~\cite{Smithviruscycle,Beltranviruscycle,Geroldviruscycle}. Besides providing mechanistic insights into the biology of infection, knowledge of virus-host interactions can point to essential events needed for virus entry, replication, or spread, which can be potential targets for the prevention, or treatment of virus-induced diseases~\cite{sadegh2020exploring}. 

\emph{In vitro} experiments based on yeast-two hybrid (Y2H), ligand-based capture MS, proximity labeling MS, and protein arrays have identified tens of thousands of virus-human protein interactions~\cite{PMID:33934825,PMID:33934830, PMID:29746851, PMID:26365680, PMID:26817613, PMID:28163258, PMID:28679763, PMID:19632888, PMID:11967329}. These interaction data are  deposited in publicly available databases including InAct~\cite{kerrien2012intact}, VirusMetha~\cite{calderone2015virusmentha}, VirusMINT~\cite{chatr2009virusmint}, and HPIDB~\cite{ammari2016hpidb}, and others. However, experimental approaches to unravel PPIs are limited by several factors, including the cost and time required, the generation, cultivation and purification of appropriate virus strains, the availability of recombinantly expressed proteins, generation of knock in or overexpression cell lines, availability of antibodies and cellular model systems. Computational approaches can assist \emph{in vitro} experimentation by providing a list of most probable interactions, which actual biological experimentation techniques can falsify or verify.

In this work, we cast the problem of predicting virus-human protein interactions as a binary classification problem and focus specifically on emerging viruses that has limited experimentally verified interaction data.

 \subsection{Key Challenges in learning to predict virus-Human PPI} 
 \mpara{Limited interaction data.} One of the main challenges in tackling the current task as a learning problem is the \emph{limited training data}. Towards predicting virus-host PPI, some known interactions of other human viruses collected from wet-lab experiments are employed as training data. The number of known PPIs is usually too small and thus, not representative enough to ensure the generalizability of trained models. In effect, the trained models might overfit the training data and would give inaccurate predictions for any given new virus.
 
\mpara{Difference to other pathogens.} A natural strategy to overcome the limitation posed by scarce virus protein interaction data is to employ transfer learning from available intra-species PPI or PPI data for other types of pathogens. This may, in its simplest fashion, not be a viable strategy as virus proteins can differ substantially from human or bacterial proteins. Typically, they are highly structurally and functionally dynamic. Virus proteins often have multiple independent functions so that they cannot be easily detected by common sequence-structure comparison~\cite{requiao2020viruses, PMID:28007618, PMID:25502394}. Besides, virus protein sequences of different species are highly diverse~\cite{eid2016denovo}.
Consequently, models trained for intra-species human PPI~\cite{li2020predicting,sun2017sequence,li2020computational,chen2019protein,sarkar2019machine} or for other pathogen-human PPI~\cite{sudhakar2020computational,mei2020silico,dick2020pipe4,li2014pathogen,guven2019interface,basit2018training} cannot be directly used to predict virus-human protein interactions.

\mpara{Limited information on structure and function of virus proteins.} While for human proteins, researchers can retrieve information from many publicly available databases to extract features related to their function, semantic annotation, domains, structure, pathway association, and intercellular localization, such information is not readily available for most virus proteins. Protein crystal structures are available for some virus proteins. However, for many, predictive structures based on the amino acid sequence must be used. Thus, for the majority of virus proteins, currently, the only reliable source of virus protein information is its amino acid sequence. 
 \emph{Learning effective representations} of the virus proteins, therefore, is an important step towards building prediction models. Heuristics such as K-mer amino acid composition
are bound to fail as it is known that virus proteins with completely different sequences might show similar interaction patterns.

\subsection{Our Contributions} In this work, we develop a machine learning model which overcomes the above limitations in two main steps, which are described below.

\mpara{Transfer Learning via Protein Sequence Representations.} Though the training data on interactions as well as the input information on protein features are limited, a large number of unannotated protein sequences are available in public databases like UniProt. Inspired by advancements in Natural Language Processing, Alley et al.~\cite{alley2019unified} trained a deep learning model on more than 24 million protein sequences to extract statistically meaningful representations. These representations have been shown to advance the state-of-the-art in protein structure and function prediction tasks. Rather than using hand-crafted protein sequence features, we use the pre-trained model by~\cite{alley2019unified} (referred to as \textsc{UniRep})  to extract protein representations. The idea here is to exploit transfer learning from several million sequences to our scant training data.  

\mpara{Incorporating Domain Information.} We further fine-tune \textsc{UniRep}'s globally trained protein representations using a simple neural network whose parameters are learned using a multitask objective. In particular, besides the main task, our model is additionally regularized by another objective, namely predicting interactions among human proteins. The additional objective allows us to encode (human) protein similarities dictated by their interaction patterns. The rationale behind encoding such knowledge in the learnt representation is that the human proteins sharing similar biological properties and functions would also exhibit similar interacting patterns with viral proteins. Using a simpler model and an additional side task helps us overcome overfitting, which is usually associated with models trained with small amounts of training data. 

We refer to our model as \textsc{MultiTask Transfer (MTT)} and is further illustrated in Section~\ref{secModel}. To sum up, we make the following contributions.

\begin{itemize}
    \item We propose a new model that employs a transfer learning-based approach to first obtain the statistically rich protein representations and then further refines them using a multitask objective. 
    \item We evaluated our approach on several benchmark datasets of different types for virus-human and bacteria-human protein interaction prediction. Our experimental results (c.f. Section \ref{secResult}) show that \textsc{MTT} outperforms several baselines even on datasets with rich feature information.
    \item Experimental results on the \sarb virus receptor shows that our model can help researchers to reduce the search space for yet unknown virus receptors effectively.
    \item We release our code for reproducibility and further development at~\href{https://git.l3s.uni-hannover.de/dong/multitask-transfer}{\url{https://git.l3s.uni-hannover.de/dong/multitask-transfer}}.
\end{itemize}

\section{Related work}
\label{secRelatedWork}
Existing work mainly casts PPI prediction task as a supervised machine learning problem. Nevertheless, the information about non-interacting protein pairs is usually not available in public databases. Therefore, researchers can only either adapt models to learn from only positive samples or employ certain negative sampling strategy to generate negative examples for training data. Since the quality and quantity of the generated negative samples would significantly affect the outcome of the learned models, the authors in~\cite{nouretdinov2012determining,li2014pathogen,nourani2016computational} proposed models that only learned from the available known positive interactions. Nourani et al.~\cite{nourani2016computational} and Li et al.~\cite{li2014pathogen} treated the virus-human PPI problem as a matrix completion problem in which the goal was to predict the missing entries in the interaction matrix. Nouretdinov et al.~\cite{nouretdinov2012determining} use a conformal method to calculate p-values/confidence level related to the hypothesis that two proteins interact based on similarity measures between proteins.

Another line of work which casts the problem as a binary classification task focussed on proposing new negative sampling techniques. For instance, Eid et al~\cite{eid2016denovo} proposed Denovo - a negative sampling technique based on virus sequence dissimilarity. Mei et al.~\cite{mei2015novel} proposed a negative sampling technique based on one class SVM.  Basit et al.~\cite{basit2018training} offered a modification to the Denovo technique by assigning sample weights to negative examples inversely proportional to their similarity to known positive examples during training. 

Dick et al. \cite{dick2020pipe4} utilizes the interaction pattern from intra-species PPI networks to predict the inter-species PPI between human-HIV-1 virus and human. Though the results are promising, this cannot be directly applied to completely new viruses where information about closely-related species is not available or to viruses whose intra-species PPI information is not available.  

The works presented in ~\cite{cui2012prediction, kim2017improved, loaiza2020predhpi, zhou2018generalized,ma2020seq,deng2020predict,dey2020machine} employed different feature extraction strategies to represent a virus-human protein pair as a fixed-length vector of features extracted from their protein sequences. Instead of hard-coding sequence feature, Yang et al.~\cite{yang2020prediction} and Lanchantin et al.~\cite{lanchantin2020transfer} proposed embedding models to learn the virus and human proteins' feature representations from their sequences. However, their training data was  limited to around 500,000 protein sequences.  Though not very common, other types of information/features were also used in some proposed models besides sequence-based features. Those include protein functional information (or GO annotation) as in~\cite{wang2020prediction}, proteins domain-domain associations information as in~\cite{barman2014prediction}, protein structure information as in~\cite{lasso2019structure,guven2019interface}, and the disease phenotype of clinical symptoms as in~\cite{wang2020prediction}. One limitation of these approaches is that they cannot be generalized to novel viruses where such kind of information is not available.

Among the network-based approaches, Liu et al. and Wang et al.~\cite{liu2019predicting,wang2020network} constructed heterogeneous networks to compute virus and human proteins features. Nodes of the same type were connected by either weighted edges based on their sequence similarity or a combination of sequence similarity and Gaussian Interaction Profile kernel similarity. Deng et al.~\cite{deng2020predict} proposed a deep-learning-based model with a complex architecture of convolutional and LSTM layers to learn the hidden representation of virus and human proteins from their input sequence features along with the classification problem. Despite the promising performance, those studies still have the limitation posed by hand-crafted protein features.

 \section{Method}
\label{secModel}
We first provide a formal problem statement.

\mpara{Problem Statement.} We are given protein sequences corresponding to infectious viruses and their known interactions with human proteins. Given a completely new (novel) virus, its set of protein(s) $V$ along with its (their) sequence(s), we are interested in predicting potential interactions between $V$ and the human proteins.

We cast the above problem as that of binary classification. The positive samples consist of pairs of virus and human proteins whose interaction has been verified experimentally. All other pairs are considered to be non-interacting and constitute the negative samples. In Section~\ref{secSetup}, we add details on positive and negative samples corresponding to each dataset.

{\mpara{Summary of the approach.} The schematic diagram of our proposed model is presented in Figure~\ref{fig:model}. As shown in the diagram, the input to the model is the raw human and virus protein sequences which are passed through the UniRep model to extract low dimensional vector representations of the corresponding proteins. The extracted embeddings are then passed as initialization values for the embedding layers. These representations are further fine-tuned using the Multilayer Perceptron (MLP) modules (shown in blue). The fine-tuning is performed while learning to predict an interaction between two human proteins (between proteins A and B in the figure) as well as the interaction between human and virus proteins (between proteins B and C). In the following, we describe in detail the main components of our approach.

\begin{figure}[h!]
    \centering
    \includegraphics[width=0.9\textwidth]{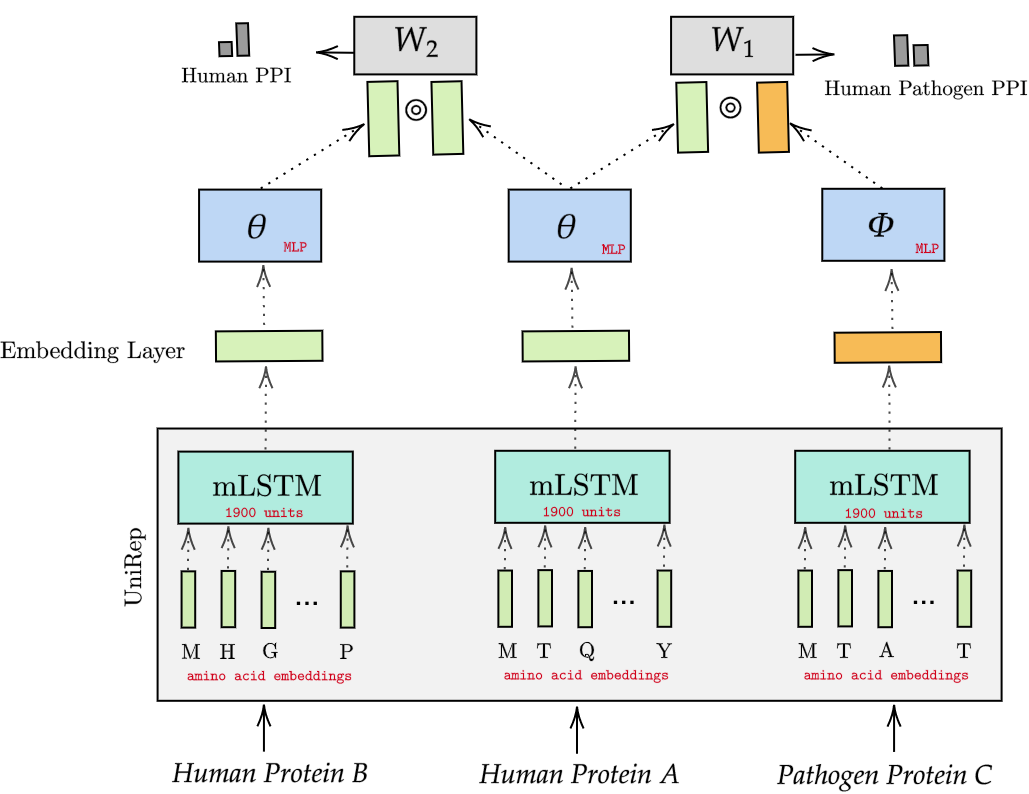}
    \caption{Our proposed \textsc{MTT} model for the virus-human PPI prediction problem. The \textsc{UniRep} embeddings are used to initialize our embedding layers which will be further fine-tuned by the two PPI prediction tasks. Sharing representation for human proteins further enables us to transfer the knowledge learned from the human PPI network to inform our virus-host PPI prediction task.}
    \label{fig:model}
\end{figure}

\subsection{Extracting protein representations}
\mpara{Significance of using protein sequence as input.} We note that the protein sequence determines the protein's structural conformation (fold), which further determines its function and its interaction pattern with other proteins. 
However, the underlying mechanism of the sequence-to-structure matching process is very complex and cannot be easily specified by hand-crafted rules. Therefore, rather than using hand-crafted features extracted from amino acid sequences, we employ the pre-trained  \textsc{UniRep} model \cite{alley2019unified} to generate latent representations or protein embeddings. The protein representations extracted from  \textsc{UniRep} model are empirically shown to preserve fundamental properties of the proteins and are hypothesized to be statistically more robust and generalizable than hand-crafted sequence features. 

\mpara{ \textsc{UniRep} for extracting sequence representations.} In particular, \textsc{UniRep} consists of an embedding layer that serves as a lookup table for each amino acid representation. Each amino acid is represented as an embedding vector of 10 dimensions. Each input protein sequence of length N will be denoted as a two-dimensional matrix of size Nx10. That two-dimensional matrix will then feed as input to a Multiplicative Long Short Term Memory (mLSTM) network of 1900 units. The 1900 dimension is selected experimentally from a pool of architectures that require different numbers of parameters as described in~\cite{biswas2020principles}, namely, a 1900-dimensional single layer multiplicative LSTM ($\sim$18.2 million parameters), a 4-layer stacked mLSTM of 256 dimensions per layer ($\sim$1.8 million parameters), and a
4-layer stacked mLSTM with 64 dimensions per layer ($\sim$0.15 million parameters). The output from mLSTM is a 1900 dimensional embedding vector that serves as the pre-trained protein embedding for the input protein sequence. We use the calculated pre-trained virus and human protein embeddings to initialize our embedding layers. The two supervised PPI prediction tasks will further fine-tune those embeddings during training.

\subsection{Learning framework}
We further fine-tune these representations by training two simple neural networks (single layer MLP with ReLu activation) using an additional objective of predicting human PPI in addition to the main task. More precisely, the \textsc{UniRep} representations will be passed through one hidden layer MLPs with ReLU activations to extract the latent representations. Let $\mathbf{X}$ denote the embedding lookup matrix. The $i$th row corresponds to the embedding vector of node $i$. The final output from MLP layers for an input $v$ is then given by $
\mathbf{hid}(v) = \operatorname{MLP}(\mathbf{X}(v))$.
To predict the likelihood of interaction between a pair $(v_1,v_2)$ we first perform an element-wise product of the corresponding hidden vectors (output of MLPs) and pass it through a linear layer followed by sigmoid activation. 
In the following we provide a detailed description of our multi-task objective.

\subsubsection{Training using a Multi-Task Objective}
 Let $\Theta,\Phi$ denote the set of learnable parameters corresponding to fine-tuning components (as shown in Figure~\ref{fig:model} in green and blue boxes), i.e., the Multilayer Perceptrons (MLP) corresponding to the virus and human proteins, respectively. Let $\mathbf{W}_1,\mathbf{W}_2$ denote the two learnable weight matrices (parameters) for the linear layers (as depicted in gray boxes in the Figure). We use $VH$, and $HH$ to denote the training set of virus-human, human-human PPI, correspondingly. We use binary cross entropy loss for predicting virus-human PPI predictions, as given below:
\begin{equation}
    \mathcal{L}_1 = \sum_{(v,h) \in VH}-z_{vh}\log y_{vh}(\Theta,\Phi, \mathbf{W}_1)
    -(1-z_{vh})\log(1- y_{vh}(\Theta,\Phi,\mathbf{W}_1)),
\end{equation}
where variables $z_{vh}$ is the corresponding binary target variable  and $y_{vh}$ is the predicted likelihood of observing virus-human protein interaction, i.e.,
\begin{align}
\label{eq:pred}
    y_{vh}(\Theta,\Phi,\mathbf{W}_1)= & \sigma((\mathbf{hid}(v) \odot \mathbf{hid}(h))\mathbf{W}_1),
\end{align}
where $\sigma(x)={1/1+e^{-x}}$ is the sigmoid activation and $\odot$ denotes the element-wise product.

For human PPI, we predict the confidence score of observing an interaction between two human proteins. More specifically, we directly predict $z_{hh'}$ - the normalized confidence scores for interaction between two human proteins as collected from STRING~\cite{szklarczyk2015string} database. Predicting the normalized confidence scores helps us overcome the issues with defining negative interactions. We use mean square error loss to compute the loss for the human PPI prediction task as below where $y_{hh'}$ is computed similar to \eqref{eq:pred} for  human proteins and $N$ is the number of $(h,h')$ pairs.
\begin{align}
    \mathcal{L}_2 = \frac{1}{N} &\sum_{(h,h') \in HH}(y_{hh'}(\Theta,\mathbf{W}_2) - z_{hh'})^{2} 
\end{align}

We use a linear combination of the two loss functions to train our model.
\begin{equation}
   \mathcal{L} =   \mathcal{L}_1 + \alpha \cdot   \mathcal{L}_2
\end{equation}

where $\alpha$ is the human PPI weight factor.

\section{Data Description and Experimental set up}
\label{secSetup}

We commence by describing the 13 datasets used in this work to evaluate our approach.
\subsection{Benchmark datasets}
{
\subsubsection{The realistic host cell-virus testing datasets}
\textbf{The \novelh and \novele datasets.}
\label{new data}
We retrieve the curated  or  experimentally  verified PPIs between virus and human from four databases: APID~\cite{alonso2016apid}, IntAct~\cite{kerrien2012intact}, VirusMetha~\cite{calderone2015virusmentha}, and UniProt~\cite{uniprot2015uniprot} using  the  PSICQUIC  web  service~\cite{aranda2011psicquic}. In total, there are 11,491 known PPIs between 246 viruses and humans. From this source of data, we generate new  training and testing data for the two viruses: the human H1N1 Influenza virus and Ebola virus. We name the two datasets \novelh and \novele according to the virus present in the testing set. The positive training data for the \novelh dataset includes PPIs between human and all viruses except H1N1. Similarly, the positive training data for the \novele dataset includes PPIs between human and all viruses except Ebola. The positive testing data for the human-H1N1 dataset contains PPIs between human and 11 H1N1 virus proteins. Likewise, the positive testing data for the human-Ebola dataset contains PPIs between human and three of the eight Ebola virus proteins (VP24, VP35, and VP40).

Negative sampling techniques such as the dissimilarity-based method~\cite{eid2016denovo}, the exclusive co-localization method~\cite{martin2005predicting,mei2013probability} are usually biased as they restrict the number of tested human proteins. It is also unrealistic for a new  virus because information about such restricted human protein set, generated from filtering criteria based on the positive instances, is typically unavailable. For those reasons, we argue that random negative sampling is the most appropriate, unbiased approach to generate negative training/testing samples. Since the exact ratio of positive:negative is unknown, we conducted experiments with different negative sample rates. In our new  virus-human PPI experiments, we try four negative sample rates: [1,2,5,10]. In addition, to reduce the bias of negative samples, the negative sampling in the training and testing set is repeated ten times. In the end, for each dataset, we test each method with 4x4x10 = 160 different combinations of negative training and negative testing sets (with fixed positive training and test samples). The statistics for our new  testing datasets are given in Table~\ref{tab:dataset}.

\begin{table}[h!]
     \centering
     \caption{The virus-human PPI realistic benchmark datasets' statistics. $|E^{+}|$ and $|E^{-}|$ refer to the number of positive and negative interactions, respectively. $|V^{h}|$ and $|V^{v}|$ are the number of human proteins and virus proteins.}
     \begin{tabular}{lrrrrcrrrr}
 
          & \multicolumn{4}{c}{\textsc{Training data}} && \multicolumn{4}{c}{\textsc{Testing data}}\\
        
         & $|E^{+}|$& $|E^{-}|$ & $|V^h|$ & $|V^v|$ & & $|E^{+}|$& $|E^{-}|$ & $|V^h|$ & $|V^v|$ \\
         \midrule
        \novelh & $10,858$ & $varies$ &  $7,636$ & $641$  & &$381$ & $varies$ & $622$ & $11$ \\ 
         \novele & $11,341$ & $varies$ & $7,816$ & $659$ & & $150$ & $varies$ & $290$ & $3$ \\
         \midrule
         \zhouh & $10,858$ & $10,858$ &  $7,636$ & $641$  & &$381$ & $381$ & $622$ & $11$ \\ 
         \zhoue & $11,341$ & $11,341$ & $7,816$ & $659$ & & $150$ & $150$ & $290$ & $3$ \\
         \midrule
         \textsc{2697049} & $24,698$ & $246,980$ & $16,638$ & $1,066$ && $278$ & $448,651$& $16,627$ & $27$ \\
         \textsc{333761} & $23,892$ & $238,920$& $16,638$ & $1,070$ && $534$ & $132,482$& $$16,627$$ & $8$ \\ 
         \textsc{2043570} & $24,372$ & $243,720$ & $16,638$ & $1,085$ && $309$ & $66,199$& $16,627$ & $4$ \\
         \textsc{644788} & $24,825$ & $248,250$& $16,638$ & $1,090$ && $54$ & $33,200$& $16,627$ & $2$ \\ 
    \end{tabular}
    
    \label{tab:dataset}
\end{table}

\mpara{The \deepviral~\cite{wang2020prediction} Leave-One-Species-Out (LOSO) benchmark datasets.}
The data was retrieved from the HPIDB database~\cite{ammari2016hpidb} to include all \emph{Pathogen-Host} interactions that have confidence scores available and are associated with an existing virus family in the NCBI taxonomy~\cite{federhen2012ncbi}. After filtering, the dataset includes 24,678 positive interactions and 1,066 virus proteins from 14 virus families. We follow the same procedure as mentioned in~\cite{wang2020prediction} to generate the training and testing data corresponding to four virus species with taxon IDs: 644788 (Influenza A), 333761 (HPV 18), 2697049 (SARS-CoV-2), 2043570 (Zika virus). From now on, we will use the NCBI taxon ID of the virus species in the testing set as the dataset name. For each dataset, the positive testing data consists of all known interactions between the test virus and the human proteins. The negative testing data consists of all possible combinations of virus and $16,627$ human proteins in Uniprot (with a length limit of 1000 amino acids) that do not appear in the positive testing set. Similarly, the positive training data consists of all known interactions between human protein and any virus protein, except for the one which is in the testing set. The negative training data is generated randomly with the positive:negative rate of 1:10 from the pool of all possible combinations of virus and $16,627$
human proteins that do not appear in the positive training set. Statistics of the datasets are presented in table~\ref{tab:dataset}. Though performing a search on the set of $16,627$ human proteins might not be a fruitful realistic strategy, we still keep the same training and testing data as released in the \deepviral study in our experiments to have a direct and fair comparison with the \deepviral method.
}
\subsubsection{The widely used new  virus-human PPI prediction benchmarked datasets.}
The two datasets released by Zhou et al.~\cite{zhou2018generalized} are widely used by recent papers to evaluate state-of-the-art models on new  virus-human PPI prediction tasks. We refer to them as \zhouh and \zhoue where each dataset was named after the viruses in the testing sets. \zhouh and \zhoue share similar positive training and testing samples with the \novelh and \novele datasets. However, they differ in the negative training and testing samples sets. While the negative samples in \novelh and \novele were generated randomly from the pool of all possible pairs, the negative training/testing samples in \zhouh and \zhoue were generated based on the protein sequence dissimilarity score. Therefore, \zhouh and \zhoue have the limitations as mentioned in section~\ref{new data} and are not ideal for evaluating the new  virus-human PPI prediction task. 
The data statistics for these two datasets are shown in Table~\ref{tab:dataset}.

\subsubsection{The specialized testing datasets}
\textbf{The dataset with protein motif information (\textsc{Denovo SLiM}~\cite{eid2016denovo}).}
The \textsc{Denovo SLiM} dataset Virus-human PPIs were collected from VirusMentha database~\cite{calderone2015virusmentha}. The presence of Short Linear Motif (SLiM) in virus sequences was used as a criterion for data filtering. SLiMs are short, recurring patterns of protein sequences that are believed to mediate protein-protein interaction~\cite{diella2008understanding,neduva2006peptides}. Therefore, sequence motifs can be a rich feature set for virus-human PPI prediction tasks. The test set~\cite{eid2016denovo} contained 425 positives and 425 negative PPIs (Supplementary file S12 used in DeNovo's study ST6). The training data consisted of the remaining PPI records and comprised of 1590 positive and 1515 negative records for which virus SLiM sequence is known and 3430 positives and 3219 negatives without virus SLiM sequences information. \slim negative samples were also generated using the Denovo negative sampling strategy (based on sequence dissimilarity).\\\\
\textbf{The \textsc{Barman's} dataset~\cite{barman2014prediction} with protein domain information}. 
The dataset was retrieved from VirusMINT database~\cite{chatr2009virusmint}. Interacting protein pairs that did not have any ``InterPro'' domain hit were removed. In the end, the dataset contained 1,035 positives and 1,035 negative interactions between 160 virus proteins of 65 types and 667 human proteins. 5-Fold cross-validation was then employed to test each method's performance.

\subsubsection{The bacteria human PPI prediction task.} 
We evaluate our method on three datasets for three human pathogenic bacteria: \textsc{Bacillus anthracis} (B1), \textsc{Yersinia pestis} (B2),  and \textsc{Francisella  tularensis} (B3), which were shared by Fatma et al.~\cite{eid2016denovo}.

The data was first collected from  HPIDB~\cite{ammari2016hpidb}.  B1 belongs to a bacterial phylum different from that of B2 and B3, while B2 and B3 share the same class but differ in their taxonomic order. B1 has 3057 PPIs, B2 has 4020, and B3 has 1346 known PPIs. A sequence-dissimilarity-based negative sampling method was employed to generate negative samples. For each bacteria protein, ten negative samples were generated randomly. Each of the bacteria was then set aside for testing, while the interactions from the other two bacteria were used for training. For simplicity, we use the name of the bacteria in the testing set as the name of the dataset. The statistics for those three datasets are presented in table~\ref{tab:bacteria_stat}.

\begin{table}[h!]
     \centering
     \caption{Our bacteria-human PPI benchmark datasets' statistics. $|E^{+}|$ and $|E^{-}|$ refer to the number of positive and negative interactions, respectively. $|V^{h}|$ and $|V^{b}|$ are the number of human proteins and bacteria proteins.}
     \begin{tabular}{lrrrrcrrrr}
 
          & \multicolumn{4}{c}{\textsc{Training data}} && \multicolumn{4}{c}{\textsc{Testing data}}\\
        
         & $|E^{+}|$& $|E^{-}|$ & $|V^h|$ & $|V^b|$ & & $|E^{+}|$& $|E^{-}|$ & $|V^h|$ & $|V^b|$ \\
         \midrule
         \textsc{Bacillus anthracis} & $5366$ & $15590$& $1559$ & $2674$ && $3057$ & $9440$& $944$ & $1705$ \\
         \textsc{Yersinia pestis} & $4403$ & $12880$& $1288$ & $2278$ && $4020$ & $12150$& $1215$ & $2147$ \\ 
         \textsc{Francisella  tularensis} & $7077$ & $21590$& $2159$ & $3041$ && $1346$ & $3440$& $344$ & $1023$ \\
    \end{tabular}
    
    \label{tab:bacteria_stat}
\end{table}

\subsection{Description of Compared Methods}

We compare our method with the following seven baseline methods and two simper variants of our model. 
\begin{itemize}
    \item \textsc{Generalized}~\cite{zhou2018generalized}: It is a generalized SVM model trained on hand-crafted features extracted from protein sequence for the novel virus-human PPI task. Each virus-human pair is represented as a vector of 1,175 dimensions extracted from the two protein sequences.
    \item \textsc{Hybrid}~\cite{deng2020predict}: It is a complex deep model with convolutional and LSTM layers for extracting latent representation of virus and human proteins from their input sequence features and is trained using L1 regularized Logistic regression.
    \item \textsc{doc2vec}~\cite{yang2020prediction}: It employs the doc2vec~\cite{le2014distributed} approach to generate protein embeddings from the corpus of protein sequences. A random forest model is then trained for the PPI prediction.
\item \textsc{MotifTransformer}~\cite{lanchantin2020transfer}: It is a transformer-based deep neural network that pre-trains protein sequence representations using unsupervised language modeling tasks and supervised protein structure and function prediction tasks. These representations are used as input to an order-independent classifier for the PPI prediction task.
\item \textsc{DeNovo}~\cite{eid2016denovo}: This model trained an SVM classifier on a hand-crafted feature set extracted from the K-mer amino acid composition information using a novel negative sampling strategy. Each protein pair is represented as a vector of 686 dimensions.
\item \deepviral~\cite{wang2020prediction}: It is a deep learning-based method that combines information from various sources, namely, the disease phenotypes, virus taxonomic tree, protein GO annotation, and proteins sequences for intra- and inter-species PPI prediction.
\item \textsc{Barman}~\cite{barman2014prediction}: It used an SVM model trained on a feature set consisting of the protein domain-domain association and methionine, serine, and valine amino acid composition of viral proteins. 
\item 2 simpler variants of MTT: Towards ablation study, we evaluate two simpler variants: (i) \textsc{SingleTask Transfer (STT)}, which is trained on a single objective of predicting pathogen-human PPI. \stt is basically the \mtt without the human PPI prediction side task
and (ii) \textsc{Naive Baseline}, which is a  Logistic regression model using concatenated human and pathogen protein \textsc{UniRep} representations as input.
    \end{itemize}
    
    \subsection{Implementation details and parameter set up}
We use Pytorch~\cite{pytorch} to implement our model and run it on an Nvidia GTX 1080-Ti with 11GB memory. We use Adam optimizer for the model parameter optimization. For all datasets, we left out 10\% of the training data for validation and performed a grid search for the best combination of parameters on that validation set. For datasets other than \novelh and \novele, we perform parameter grid searching with the MLP hidden dimension $hid$ in [8, 16,32, 64], $\alpha$ in $[10^{-3}, 10^{-2}, 10^{-1}, 1]$, the number of $epochs$ from 0 to 200 with a step of 2 and the learning rate $lr$ in $[10^{-3}, 10^{-2}]$. For the \novelh and \novele datasets, we test each with 160 different combinations of negative training and negative testing. Therefore, we fix the hidden dimension to 16, $\alpha=10^{-3}$, $lr=10^{-3}$ and only perform grid searching on the number of epochs. The reported results for each dataset are the results corresponding to the best-performed model on the validation set.

For the \docvec model, we use the released code shared by the authors with the given parameters. For the \generalized and \denovo models, we re-implement the methods in Python using all the parameters and feature set as described in the original papers. For \barman and \deepviral, the results are taken from the original papers or calculated from the given model prediction scores.

\subsection{Evaluation metrics}
For all benchmark datasets except the case study, we report five metrics: the Area under Receiver Operating Characteristic curve (\textsc{AUC}) and the area under the precision-recall
curve (\textsc{AP}), the \textsc{Precision}, \textsc{Recall}, and \textsc{F1} scores.

For the case study, we report the topK score with $K$ from 1 to 10. TopK is equal to 1 if the human receptor for \sarb virus appears in the top $K$ proteins that have the highest scores predicted by the model and 0 otherwise.

\section{Result Analysis}
\label{secResult}
In the following four subsections, we provide a detailed comparison of \mtt with (i) methods employing hand-crafted input features, (ii) sequence embedding-based methods, (iii) an approach that uses protein domain information, (iv) simpler variants of \mtt as ablation studies respectively. All statistical test results present in this section are those from the pair-wise t-test~\cite{welch1947generalization} on the F1 scores attained from multiple runs on the same dataset.

\begin{figure}[h!]
    \centering
    \includegraphics[width=.95\textwidth]{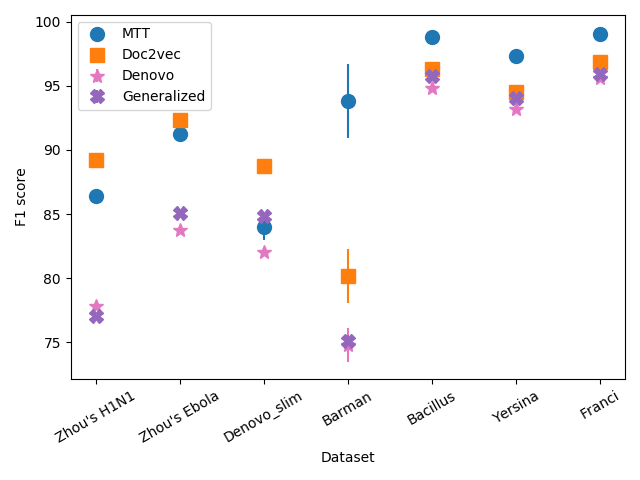}
    \caption{Comparison between \mtt and state-of-the-art methods on small testing datasets. \mtt is statistically better than \denovo in all benchmarked datasets. Compared with \generalized, \mtt has higher performance in six out of seven datasets. \mtt outperforms \docvec in four out of seven datasets.}
    \label{fig:smallSOTA}
\end{figure}

\subsection{Comparison with methods employing hand-crafted features}
\label{sec:comparehardcode}
\generalized~\cite{zhou2018generalized} and \denovo~\cite{eid2016denovo} are the two traditional methods relying on hand-crafted features extracted from the protein sequences. The number of hand-crafted features employed by \denovo and \generalized are 686 and 1,175, respectively. They both employ \textsc{SVM} for the classification task. Since \textsc{SVM} scales quadratically with the number of data points, \denovo and \generalized are not scalable to larger datasets. 

Figure~\ref{fig:smallSOTA} presents their comparison between \mtt on small testing datasets. Detailed scores are given in Table~\ref{tab:hardcode} in the Appendix. Results from the two-tailed t-test~\cite{salzberg1997comparing,kafadar1997handbook} support that \mtt significantly outperforms \denovo in all benchmarked datasets with a confidence score of at least $95\%$. Compared with \generalized, \mtt has higher performance in six out of seven datasets (except \slim). The difference is the most significant on the \barman, \zhouh, and \zhoue datasets. On \slim dataset, \mtt's F1 score is lower than \generalized and only 2\% higher than \denovo. This is expected since \slim is a specialized dataset favoring methods using local sequence motif features, which are exploited by \denovo and \generalized.

\textsc{Hybrid} is one recently proposed, deep learning-based method. Despite that, the input features are still manually extracted from the protein sequence. Since the code is not publicly available, we only have the \textsc{AUC} score corresponding to the \zhouh dataset, which is also taken from the original paper as listed in table~\ref{tab:hardcode}. Compared with \textsc{Hybrid}, \mtt has higher \textsc{AUC} score. Though comparison on the \textsc{AUC} for one dataset does not bring much insight, we include this method here for completeness.

\begin{figure}[h!]
    \centering
     \subfigure[\novele]{\includegraphics[width=.95\textwidth]{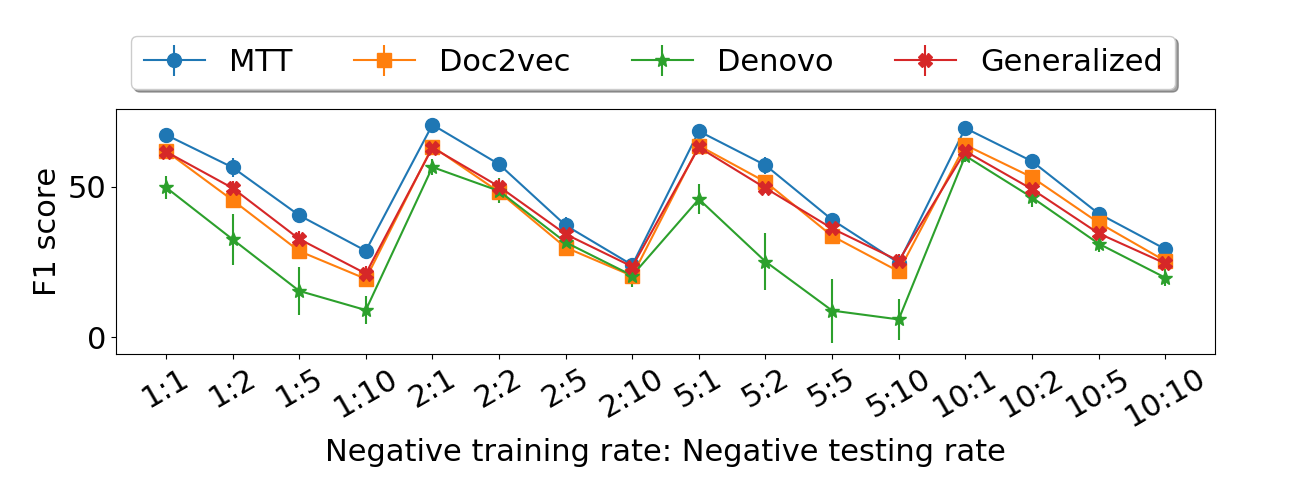}} 
     
    \subfigure[\novelh]{\includegraphics[width=.95\textwidth]{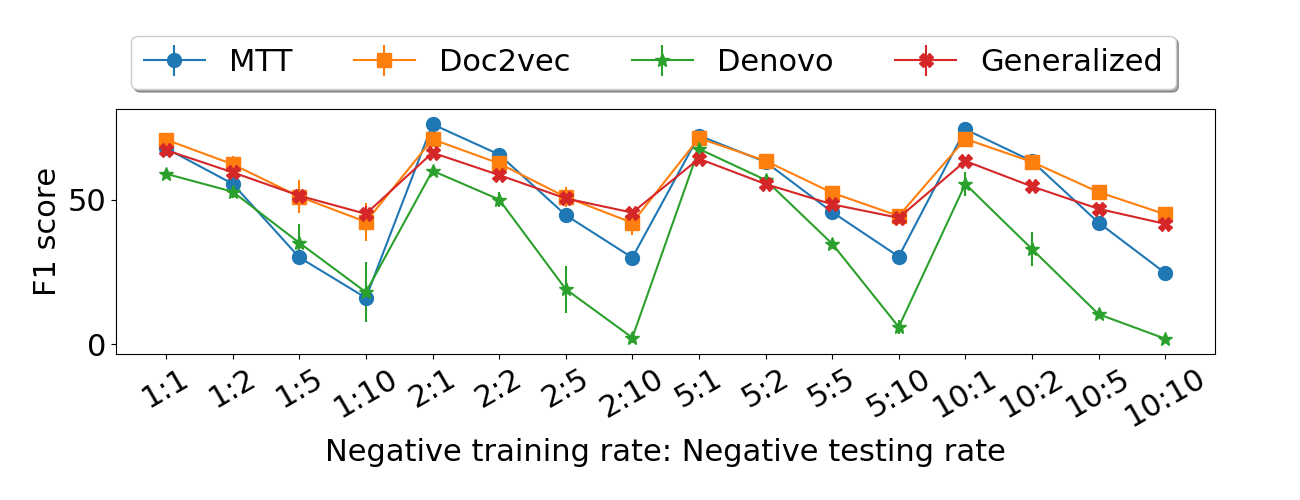}} 
   \caption{Comparison between \mtt and state-of-the-art methods on the \novele and \novelh datasets over different combinations of negative training and testing sets. \mtt is significantly better than \docvec on the \novele dataset (a), while on the \novelh dataset (b), the reverse holds true. \mtt is statistically better than \denovo on both datasets. For the \generalized model, we can only have results up to the negative training rate of 2 because, for larger negative training rates, the model took days to finish one run.}
    \label{fig:novelEmbedding}
\end{figure}
\subsection{Comparison with sequence embedding based methods}
\label{sec:comparesequence}
\docvec and \textsc{MotifTransformer} are state-of-the-art methods based on sequence embeddings or representations. \docvec utilizes the embeddings learned from the extracted k-mer features while \mtt and \motif employ the embedding directly learned from the amino acid sequences. In addition, \mtt is a multitask-based approach that incorporates additional information on human protein-protein interaction into the learning process. 

Figure~\ref{fig:novelEmbedding} shows a comparison in F1 score of \mtt and \docvec over all benchmarked datasets. Detailed scores are presented in Table~\ref{tab:embedding} in the Appendix.
Since the code for the \motif model is not publicly available, we only have the corresponding results available for the \zhouh and \zhoue datasets, which are also taken from the original paper. `-' denotes the score is not available. Compared with \motif, \mtt has a slightly worse F1 score on \zhouh and significantly better F1 score on \zhoue datasets. 

\mpara{Comparison with \docvec.} \mtt out-performs \docvec in 5 out of 9 benchmark datasets, and the performance gap is statistically significant with a p-value smaller than 0.05.  \mtt is significantly better than \docvec on the \novele dataset, while on the \novelh dataset, the reverse holds true. 
\docvec outperforms \mtt in three testing datasets whose negative samples were drawn from a sequence dissimilarity method. 
We also note that these datasets might be biased since in the ideal testing scenario, we do not have knowledge about the set of human proteins that interacted with the virus. Therefore, such dissimilarity-based negative sampling is infeasible.

\subsection{Comparison with methods that use domain information}
\barman features set is constructed from the domain-domain association and the hand-crafted feature extracted from the protein sequences. Since the protein domain information is not available for all viral proteins, the \barman method has restricted application. A comparison between \barman and \mtt is presented in table~\ref{tab:barman}. Due to data and code availability, we only have the results for the \barman model on one dataset. From reported results, we could clearly see that \mtt outperforms its competitor for a large margin in all available metrics. 

\begin{table}[h!]
\centering
     \caption{Comparison between \mtt and \barman - a method that relies on the protein domain information. Due to data and code availability issues, for the \barman method, we only have results for the \barman's dataset, which are also taken from the original paper. `$-$' indicates that the result is not available.  }
    \begin{tabular}{llccccc}
         \textsc{Model} & \textsc{AUC} & \textsc{AP} &  \textsc{Precision} &  \textsc{Recall} & \textsc{F1} \\
        \toprule
        \barman & $0.7300$ & $-$ & $-$ & $67.00$ & $69.41$ \\ 
      \mtt& $0.9804$ & $0.9802$ & $93.53$ & $94.05$ & $93.79$ \\
         
    \end{tabular}
   
    \label{tab:barman}
\end{table}

\subsection{Comparison with methods that used GO, taxonomy and phenotype information}
\deepviral exploited that disease phenotypes, the viral taxonomies, and proteins' GO annotation to enrich its protein embeddings. Table~\ref{tab:deepviral} presents a comparison between \mtt and \deepviral on the four datasets released by \deepviral's authors. The reported results on each dataset are the average after five experimental runs for \deepviral and ten experimental runs for \mtt. We observe \mtt and \stt significantly supersede their competitor regarding the averaged F1 score. The gain is more significant on smaller datasets (\textsc{644788} and \textsc{333761})

\begin{table}[h!]
\centering
\caption{Comparison with \deepviral - a method that can utilize knowledge from the disease phenotype, virus taxonomy, the human PPI network, and the protein GO annotation. Results from the pair-wise t-test indicate that \mtt is significantly better than \deepviral on three datasets (2697049, 333761, and 2043570) with a p-value smaller than $0.05$. On the $644788$ dataset, the difference is not statistically significant.}
    \begin{tabular}{llccccc}
         \textsc{Dataset} & \textsc{Model} & \textsc{AUC} & \textsc{AP} &  \textsc{Precision} &  \textsc{Recall} & \textsc{F1}\\
        \toprule
        \multirow{2}{*}{\textsc{2697049}} &\deepviral  & $0.7288$ & $0.0015$ & $0.07$ & $0.07$ & $0.07$\\
  
    & \mtt& $\textbf{0.7566}$ & $\textbf{0.0021}$ & $\textbf{0.97}$ & $\textbf{0.97}$ & $\textbf{0.97}$\\ 
      
      \midrule
      \multirow{2}{*}{\textsc{333761}} &\deepviral  & $0.8009$ & $0.0147$ & $1.72$ & $1.72$ & $1.72$ \\ 
    
      & \mtt& $\textbf{0.8160}$ & $\textbf{0.0262}$ & $\textbf{6.35}$ & $\textbf{6.35}$ & $\textbf{6.35}$\\
    
      \midrule
      \multirow{2}{*}{\textsc{2043570}} &\deepviral  & $\textbf{0.7708}$ & $\textbf{0.0116}$ & $0.52$ & $0.52$ & $0.52$\\ 
      & \mtt& $0.6956$ & $0.0096$ & $\textbf{1.89}$ & $\textbf{1.91}$ & $\textbf{1.90}$\\ 
   
      \midrule
      \multirow{2}{*}{\textsc{644788}} &\deepviral  & $0.9325$ & $\textbf{0.0357}$  & $\textbf{3.70}$ & $\textbf{3.70}$ & $3.70$\\
    & \mtt& $\textbf{0.9537}$ & $0.0302$  & $3.54$ & $\textbf{22.04}$ & $\textbf{5.46}$\\ 
    
    \end{tabular}
    
    \label{tab:deepviral}
\end{table}

\subsection{Ablation Studies}
\begin{figure}[!t]
    \centering
    \subfigure[Small testing datasets]{\includegraphics[width=.8\textwidth]{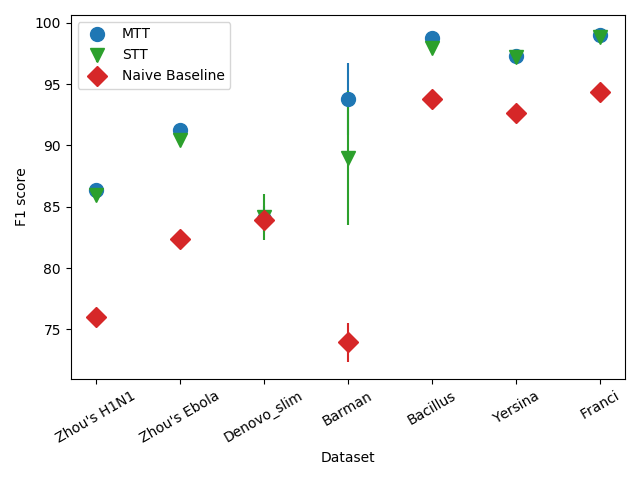}}

    \subfigure[\novelh]{\includegraphics[width=.8\textwidth]{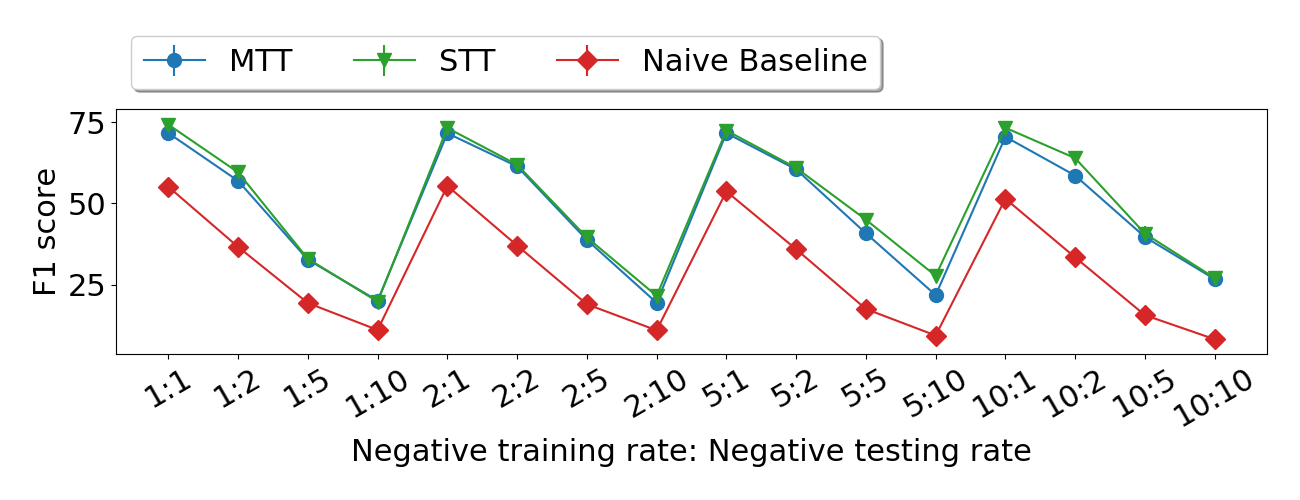}}
    \subfigure[\novele]{\includegraphics[width=.8\textwidth]{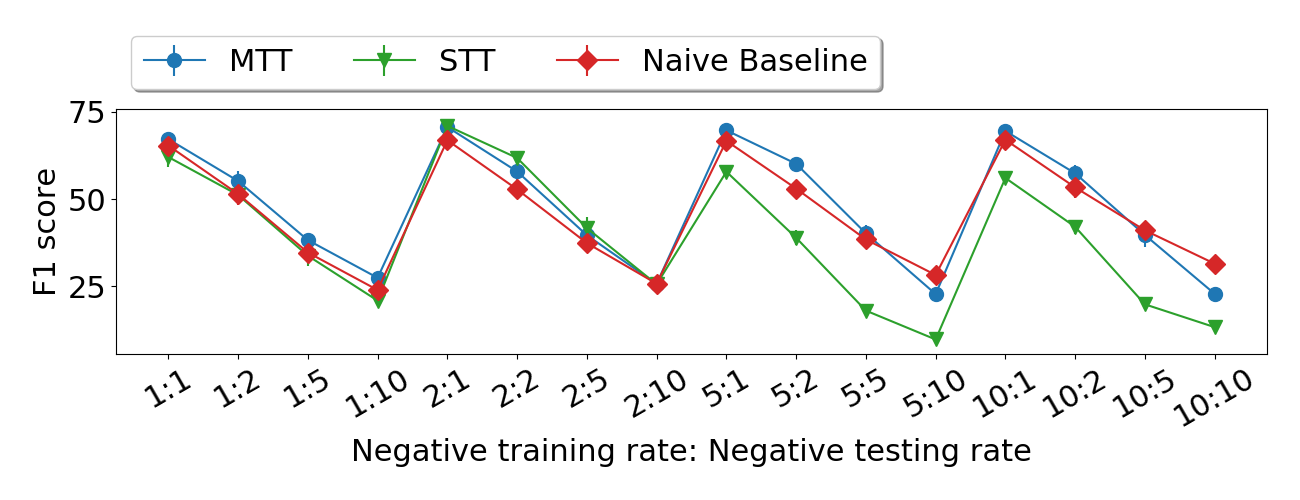}}
    \caption{Ablation study on benchmarked datasets. Compared with \stt, \mtt is statistically better in five datasets, while on the remaining four (\novelh, \slim, \textsc{Yersina}, and \textsc{Franci}), the difference is not statistically significant. \mtt is statistically better than \naive on eight out of nine datasets, while on the remaining dataset(\novele), the difference is not statistically different.}
    \label{fig:ablation}
\end{figure}
We compare our method with two of its simpler variants: the \stt and the \naive baseline models. \stt is the \mtt model without the human PPI prediction task. \naive concatenates the learned embeddings for the virus and human proteins to form the input to a Logistic Regression model. Figure~\ref{fig:ablation} presents a comparison between the F1 score of \mtt and its variants on our benchmarked datasets. Table~\ref{tab:ablation} show all reported scores over all datasets. \mtt is significantly better than \stt in five out of nine benchmarked and the four \deepviral datasets with a p-value smaller than 0.05. While in the remaining four datasets, the difference is not statistically significant. This confirms that the learned patterns from the human PPI network bring additional benefits to the virus-human PPI prediction task. 

Compare with \naive, \mtt wins in eight out of nine benchmarked and the four \deepviral datasets. On the remaining dataset (\novelh), the difference is not statistically different. \stt significantly outperforms \naive in eight out of nine datasets. This claims the effectiveness of our chosen architecture.

\section{Case study for SARS-CoV-2 binding prediction}
\label{case_study_data}
The virus binding to cells or the interaction between viral attachment proteins and host cell receptors is the first and decisive step in the virus replication cycle. Identifying the host receptor(s) for a particular virus is often fundamental in unveiling the virus pathogenesis and its species tropism. 

Here we present a case study for detecting the human protein binding partners for \sarb.
Our virus-human PPI dataset is retrieved from the InAct Molecular Interaction database~\cite{kerrien2012intact} (the latest update is 07.05.2021). We retrieve the protein sequences from Uniprot~\cite{uniprot2015uniprot}. In the next section, we describe the construction of the training and testing dataset to predict \sarb binding partners.

\subsection {Training, Validation and Test Sets for Virus-Human PPI}
The statistics for our \sarb binding prediction dataset are presented in table~\ref{tab:binding_stat}. We construct the corresponding datasets as follows.

\mpara{Training Set.} As positive interaction samples, we include in the training data only  \emph{direct} interactions between the human proteins and any virus except the \sara and \sarb. \emph{Direct} interaction requires two proteins to directly bind to each other, i.e. without an additional bridging protein. Moreover, the interacting human protein should be on the cell surface. Without loss of generality, we perform our search for the binding receptor on the set of all human proteins that have a \emph{KNOWN direct interaction} with any virus and \emph{locate} to the cell surface. Our surface human protein list consists of all reviewed Uniprot proteins that meet at least one of the following criteria: (i) appears in the human surfacetome~\cite{bausch2015mass} list or (ii) has at least one of the following GO annotations~\cite{ashburner2000gene,gene2021gene}:\{\emph{CC-plasma membrane, CC-cell junction}\}. 

The negative samples for training data contain \emph{indirect} (interactions that are not marked as direct in the database) between the human proteins and any virus except \sara and \sarb. The \emph{indirect} interactions can be a physical association (two proteins are detected in the same protein complex at the same point of time) or an association in which two proteins that may participate in the formation of one or more physical complexes without additional evidence whether the proteins are directly binding to specific members of such a complex).

\mpara{Validation and Test Sets.} As established in studies~\cite{shang2020cell, zhang2021molecular,hoffmann2020sars}, angiotensin-converting enzyme 2 (ACE2) is the human receptor for both \sara \cite{PMID:14647384} and \sarb viruses~\cite{hoffmann2020sars}. The positive validation and testing set consist of interaction between the known human receptor (ACE2) and the corresponding spike proteins of \sara and \sarb, respectively. Our negative validation and testing set encapsulate of all possible combinations the two viral spike proteins and $52$ human proteins that meet our filtering criteria.

\subsection{The intra human PPI for the Side Task}
Since we are interested in only the direct interaction between virus and human proteins, we also customize our intra human PPI training set. Our intra human PPI dataset is also retrieved from the InAct~\cite{kerrien2012intact} database (the latest update is 07.05.2021). We retain only interactions between two human proteins that appear in the virus-human PPI dataset constructed above. The confidence scores are normalized into the [0,1] ranges. All confidence scores corresponding to ``indirect'' interactions are set to 0. In the end, our intra-human PPI training set consists of 96,458 interactions between 5,563 human proteins.

\begin{table}[h!]
     \centering
     \caption{The case study statistics. $|E^{+}|$ and $|E^{-}|$ refer to the number of positive and negative interactions, respectively. $|V^{h}|$ and $|V^{v}|$ are the number of human proteins and virus proteins.}
     \begin{tabular}{rrcrrcrrcrrcc}
        &&& \multicolumn{2}{c}{\textsc{Training}} && \multicolumn{2}{c}{\textsc{Validation}} &&\multicolumn{2}{c}{\textsc{Testing}} && \textsc{Human PPI}\\
         $|V^h|$ & $|V^v|$ && $|E^{+}|$& $|E^{-}|$ && $|E^{+}|$& $|E^{-}|$ && $|E^{+}|$& $|E^{-}|$ && $|E|$  \\
         \midrule
          $5,563$ & $834$ && $554$ & $17,418$ && $1$ & $51$ && $1$ & $51$ && $96,459$\\ 

    \end{tabular}
    
    \label{tab:binding_stat}
\end{table}

\subsection{Results}
\begin{figure}[!t]
    \centering
    \includegraphics[width=.9\textwidth]{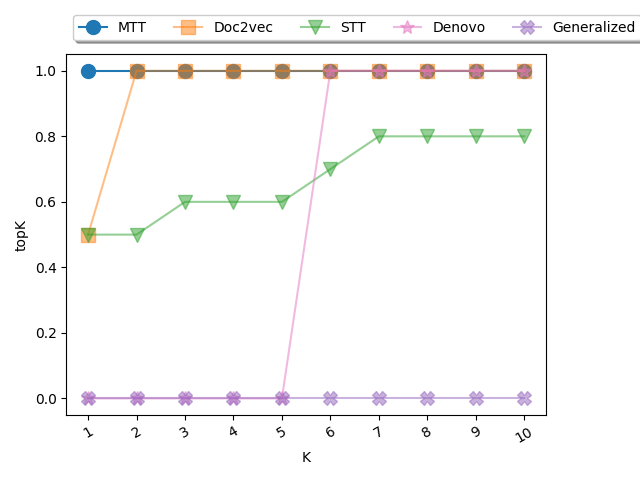}
    \caption{Case study results for benchmarked methods. $topK=1$ if the \sarb virus receptor appear in the top K proteins that have highest scores predicted by the model and $topK=0$ otherwise. The reported results are the averages after 10 runs.}
    \label{fig:casestudy}
\end{figure}
\begin{figure}[!t]
    \centering
    \includegraphics[width=.9\textwidth]{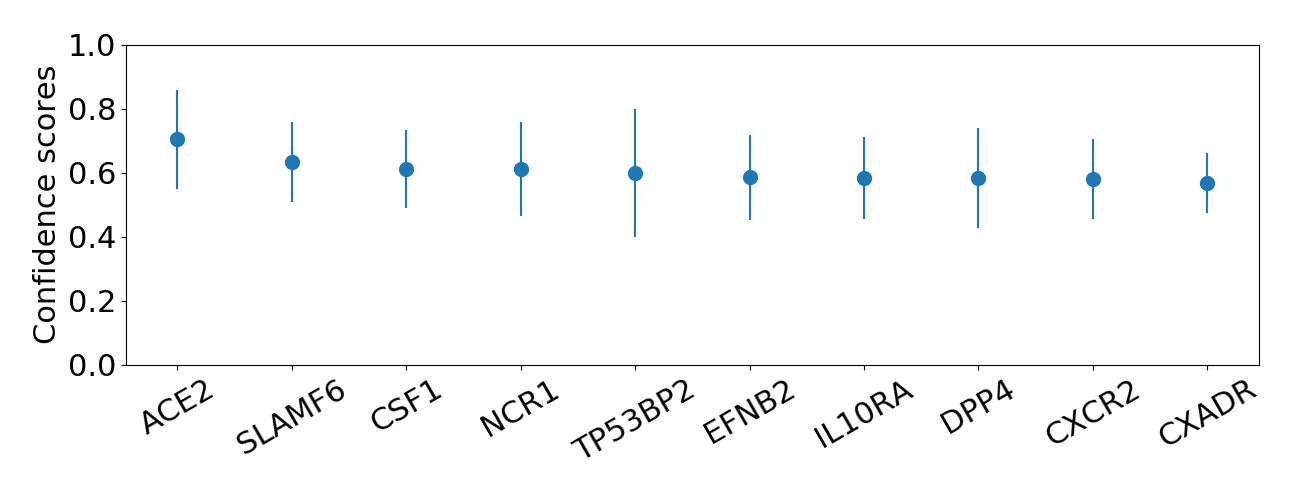}
    \caption{The top 10 predictions made by the \mtt model. The bars represent the average confidence scores after 10 experimental runs while the lines represent the standard deviation.}
    \label{fig:mttTop}
\end{figure}
\begin{figure}[!t]
    \centering
    \includegraphics[width=.9\textwidth]{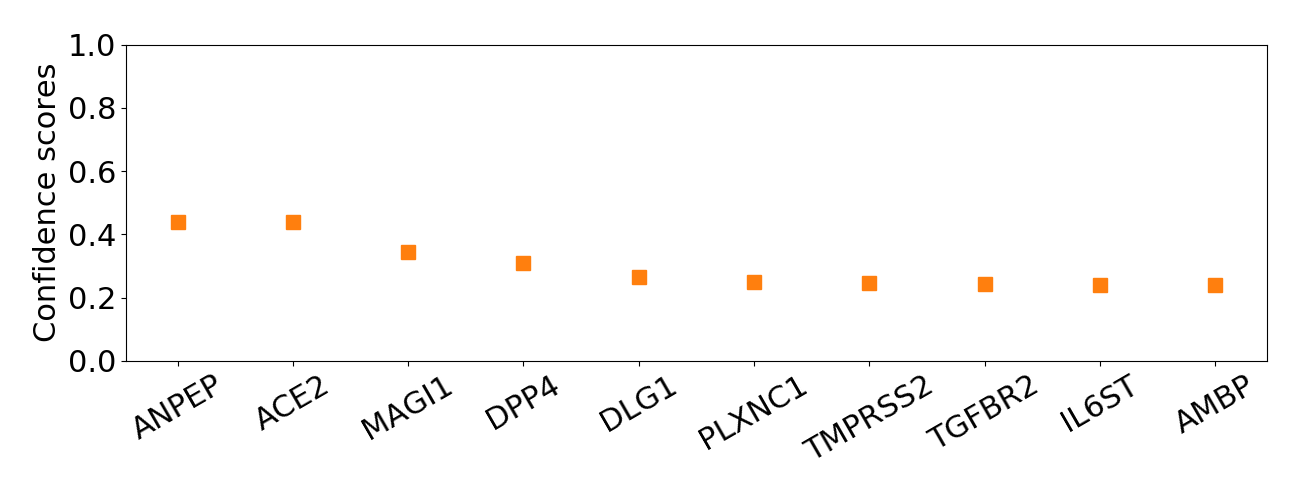}
    \caption{The top 10 predictions made by the \docvec model. The bars represent the average confidence scores after 10 experimental runs while the lines represent the standard deviation.}
    \label{fig:doc2vecTop}
\end{figure}
Finally, we here evaluate the prediction methods on how effective they are in ranking human protein candidates for binding to an emerging virus envelope protein. 
Figure~\ref{fig:casestudy} presents the methods' performance after ten runs on the case study dataset. 
TopK is equal to 1 if the true human receptor appears in the top K proteins that correspond to the highest predicted scores by the model and is equal to 0 otherwise. The reported scores plotted in Figure~\ref{fig:casestudy} are the average after ten experimental runs with random initialization.

Using this method we find that ACE2, the only \sarb receptor proven in \textit{in vivo} and \textit{in vitro} studies\cite{hoffmann2020sars, PMID:32380511, PMID:32839612}, consistently appears as the highest ranked prediction of \mtt in each of the ten experimental runs. We observe a significant difference between the highest ranked performance of \mtt and its competitors. The performance gain shown by \mtt over \stt is quite substantial after ten runs and supports the superiority of our multitask framework. The next highest nine hits presented in both models have not been shown to interact with SARS-CoV-2 in \textit{in vitro} studies. Interestingly, dipeptidyl peptidase 4 (DDP4), a receptor for another betacoronavirus MERS-CoV \cite{wang2013structure} also scored highly in the MTT method. However, although \textit{in silico} analysis has speculated a possible interaction~\cite{vankadari2020emerging}, it is yet to be shown experimentally. Similarly, the serine protease TMPRSS2, which is required for SARS-CoV-2 S protein priming during entry~\cite{hoffmann2020sars}, appeared in position 7 using the Doc2vec model. Finally, aminopeptidase N (ANPEP) the receptor for the common cold coronavirus 229E appeared as first hit in the Doc2vec model~\cite{PMID:1350662}.

In Figures \ref{fig:mttTop} and \ref{fig:doc2vecTop}, we plot the average confidence scores (corresponding to predicted interaction probability) corresponding to top 10 predictions of \mtt and \docvec models. Specifically, the proteins are ranked based on the average (over 10 runs) confidence scores as predicted by the two models. While for \mtt, the receptor ACE2 always occurs at the top of the list with average confidence score of more than 0.70 (which is more than 11\% higher than the confidence score assigned to the second hit), \docvec assigns it a score of less than 0.44 where ACE2 is ranked 2nd based on average scores. Moreover, there is negligible difference between the prediction scores for ACE2 and the first predicted hit ANPEP in case of \docvec. 

These results indicate that \mtt can provide high-quality prediction results and can help biologists to restrict the search space for the virus interaction partner effectively. This case study showcases the effectiveness of our method in solving virus-human PPI prediction problem and aims to convince biologists of the potential application of our prediction framework.

\section{Conclusion}
\label{secConclusion}
We presented a thorough overview of state-of-the-art models and their limitations for the task of virus-human PPI prediction. Our proposed approach exploits powerful statistical protein representations derived from a corpus of around 24 Million protein sequences in a multitask framework. Noting the fact that virus proteins tend to mimic human proteins towards interacting with the host proteins, we use the prediction of human PPI as a side task to regularize our model and improve generalization. The comparison of our method with a variety of state-of-the-art models on several datasets showcase the superiority of our approach. Ablation study results suggest that the human PPI prediction side task brings additional benefits and helps boost the model performance. A case study on the interaction of the \sarb virus spike protein and its human receptor indicates that our model can be used as an effective tool to reduce the search space for evaluating host protein candidates as interacting partners for emerging viruses.
In future work, we will enhance our multitask approach by incorporating more domain information including structural protein prediction tools \cite{PMID:34265844} as well as exploiting more complex multitask model architectures.

\section*{List of abbreviations}
\begin{table}[ht!]
    \flushleft
    \begin{tabular}{ll}
\textbf{ACE2}& angiotensin-converting enzyme 2\\
\textbf{ANPEP}& aminopeptidase N\\
\textbf{AP}& the area under the precision-recall curve\\
\textbf{AUC}& the area under Receiver Operating Characteristic curve\\
\textbf{DDP4}& dipeptidyl peptidase 4\\
\textbf{HH}& human-human protein-protein interaction training set\\
\textbf{LOSO}& Leave-One-Species-Out\\
\textbf{LSTM}& Long Short Term Memory\\
\textbf{mLSTM}& Multiplicative Long Short Term Memory\\
\textbf{MLP}& Multilayer Perceptrons\\
\textbf{MTT}& Multitask Transfer\\
\textbf{PPI}& protein-protein interaction\\
\textbf{SLiM}& Short Linear Motif\\
\textbf{STT}& Single task transfer\\
\textbf{VH}& virus-human protein-protein interaction training set\\
\textbf{Y2H}& yeast-two hybrid\\
\textbf{644788}& The Influenza A virus taxon ID\\
\textbf{33761}& The HPV 18 virus taxon ID\\
\textbf{697049}& The SARS-CoV-2 virus taxon ID\\
\textbf{043570}& The Zika virus taxon ID\\

    \end{tabular}
    \label{tab:abbreviation}
\end{table}

\section*{Declarations}
\begin{backmatter}

\section*{Ethics approval and consent to participate}
Not applicable

\section*{Consent for publication}
Not applicable

\section*{Availability of data and materials}
All the code and data used in this study is publicly available at ~\href{https://git.l3s.uni-hannover.de/dong/multitask-transfer}{\url{https://git.l3s.uni-hannover.de/dong/multitask-transfer}}

\section*{Competing interests}
The authors declare that they have no competing interests.

\section*{Funding}
N.D is funded by VolkswagenStiftung's initiative "Niedersächsisches Vorab" (grant no.11-76251-99-3/19 (ZN3434)). G.B and G.G are supported by the Ministry of Lower Saxony (MWK, project 76251-99 awarded to G.G.). M.K is supported the Federal Ministry of Education and Research (BMBF), Germany under the project LeibnizKILabor (grant no. 01DD20003).

The funding bodies did not play any role in the design of the study, collection, analysis, interpretation of data, and in writing the manuscript.

\section*{Authors' contributions}
N.D designed the study, collected the data, implemented the models and analyzed the results. G.B and G.G qualitatively validated the design and results of the case study. M.K designed and supervised the study as well as analyzed the results. All authors wrote the manuscript. All authors have read and approved the final manuscript.

\section*{Acknowledgements}
A preliminary version of this work~\cite{dong2021multitask} was presented at the ICLR Workshop on AI for Public Health 2021.

\bibliographystyle{bmc-mathphys} 
\bibliography{bmc_article}      %


\begin{thebibliography}{80}
\ifx \bisbn   \undefined \def \bisbn  #1{ISBN #1}\fi
\ifx \binits  \undefined \def \binits#1{#1}\fi
\ifx \bauthor  \undefined \def \bauthor#1{#1}\fi
\ifx \batitle  \undefined \def \batitle#1{#1}\fi
\ifx \bjtitle  \undefined \def \bjtitle#1{#1}\fi
\ifx \bvolume  \undefined \def \bvolume#1{\textbf{#1}}\fi
\ifx \byear  \undefined \def \byear#1{#1}\fi
\ifx \bissue  \undefined \def \bissue#1{#1}\fi
\ifx \bfpage  \undefined \def \bfpage#1{#1}\fi
\ifx \blpage  \undefined \def \blpage #1{#1}\fi
\ifx \burl  \undefined \def \burl#1{\textsf{#1}}\fi
\ifx \doiurl  \undefined \def \doiurl#1{\textsf{#1}}\fi
\ifx \betal  \undefined \def \betal{\textit{et al.}}\fi
\ifx \binstitute  \undefined \def \binstitute#1{#1}\fi
\ifx \binstitutionaled  \undefined \def \binstitutionaled#1{#1}\fi
\ifx \bctitle  \undefined \def \bctitle#1{#1}\fi
\ifx \beditor  \undefined \def \beditor#1{#1}\fi
\ifx \bpublisher  \undefined \def \bpublisher#1{#1}\fi
\ifx \bbtitle  \undefined \def \bbtitle#1{#1}\fi
\ifx \bedition  \undefined \def \bedition#1{#1}\fi
\ifx \bseriesno  \undefined \def \bseriesno#1{#1}\fi
\ifx \blocation  \undefined \def \blocation#1{#1}\fi
\ifx \bsertitle  \undefined \def \bsertitle#1{#1}\fi
\ifx \bsnm \undefined \def \bsnm#1{#1}\fi
\ifx \bsuffix \undefined \def \bsuffix#1{#1}\fi
\ifx \bparticle \undefined \def \bparticle#1{#1}\fi
\ifx \barticle \undefined \def \barticle#1{#1}\fi
\ifx \bconfdate \undefined \def \bconfdate #1{#1}\fi
\ifx \botherref \undefined \def \botherref #1{#1}\fi
\ifx \url \undefined \def \url#1{\textsf{#1}}\fi
\ifx \bchapter \undefined \def \bchapter#1{#1}\fi
\ifx \bbook \undefined \def \bbook#1{#1}\fi
\ifx \bcomment \undefined \def \bcomment#1{#1}\fi
\ifx \oauthor \undefined \def \oauthor#1{#1}\fi
\ifx \citeauthoryear \undefined \def \citeauthoryear#1{#1}\fi
\ifx \endbibitem  \undefined \def \endbibitem {}\fi
\ifx \bconflocation  \undefined \def \bconflocation#1{#1}\fi
\ifx \arxivurl  \undefined \def \arxivurl#1{\textsf{#1}}\fi
\csname PreBibitemsHook\endcsname

\bibitem{socialcostsCOVID-19}
\begin{barticle}
\bauthor{\bsnm{Petersen}, \binits{E.}},
\bauthor{\bsnm{Koopmans}, \binits{M.}},
\bauthor{\bsnm{Go}, \binits{U.}},
\bauthor{\bsnm{Hamer}, \binits{H.H.}},
\bauthor{\bsnm{Petrosillo}, \binits{N.}},
\bauthor{\bsnm{Castelli}, \binits{F.}},
\bauthor{\bsnm{Storgaard}, \binits{M.}},
\bauthor{\bsnm{Al~Khalili}, \binits{S.}},
\bauthor{\bsnm{Simonsen}, \binits{L.}}:
\batitle{Comparing sars-cov-2 with sars-cov and influenza pandemics}.
\bjtitle{The Lancet Infectious Diseases}
\bvolume{20}(\bissue{9}),
\bfpage{238}--\blpage{2244}
(\byear{2020})
\end{barticle}
\endbibitem

\bibitem{Smithviruscycle}
\begin{barticle}
\bauthor{\bsnm{Smith}, \binits{G.A.}},
\bauthor{\bsnm{Enquist}, \binits{L.W.}}:
\batitle{Break ins and break outs: viral interactions with the cytoskeleton of
  mammalian cells}.
\bjtitle{Annual Review of cell and developmental biology}
\bvolume{18},
\bfpage{135}--\blpage{61}
(\byear{2002})
\end{barticle}
\endbibitem

\bibitem{Beltranviruscycle}
\begin{barticle}
\bauthor{\bsnm{Beltran}, \binits{P.M.J.}},
\bauthor{\bsnm{Cook}, \binits{K.C.}},
\bauthor{\bsnm{Cristea}, \binits{I.M.}}:
\batitle{Exploring and exploiting proteome organization during viral
  infection}.
\bjtitle{Journal of Virology}
\bvolume{91}(\bissue{18}),
\bfpage{00268}--\blpage{17}
(\byear{2017})
\end{barticle}
\endbibitem

\bibitem{Geroldviruscycle}
\begin{barticle}
\bauthor{\bsnm{Gerold}, \binits{G.}},
\bauthor{\bsnm{Bruening}, \binits{J.}},
\bauthor{\bsnm{Weigel}, \binits{B.}},
\bauthor{\bsnm{Pietschmann}, \binits{T.}}:
\batitle{Protein interactions during the flavivirus and hepacivirus life
  cycle}.
\bjtitle{Molecular and Cellular Proteomics}
\bvolume{16}(\bissue{4 suppl 1}),
\bfpage{75}--\blpage{91}
(\byear{2017})
\end{barticle}
\endbibitem

\bibitem{sadegh2020exploring}
\begin{barticle}
\bauthor{\bsnm{Sadegh}, \binits{S.}},
\bauthor{\bsnm{Matschinske}, \binits{J.}},
\bauthor{\bsnm{Blumenthal}, \binits{D.B.}},
\bauthor{\bsnm{Galindez}, \binits{G.}},
\bauthor{\bsnm{Kacprowski}, \binits{T.}},
\bauthor{\bsnm{List}, \binits{M.}},
\bauthor{\bsnm{Nasirigerdeh}, \binits{R.}},
\bauthor{\bsnm{Oubounyt}, \binits{M.}},
\bauthor{\bsnm{Pichlmair}, \binits{A.}},
\bauthor{\bsnm{Rose}, \binits{T.D.}}, \betal:
\batitle{Exploring the sars-cov-2 virus-host-drug interactome for drug
  repurposing}.
\bjtitle{Nature communications}
\bvolume{11}(\bissue{1}),
\bfpage{1}--\blpage{9}
(\byear{2020})
\end{barticle}
\endbibitem

\bibitem{PMID:33934825}
\begin{barticle}
\bauthor{\bsnm{Wendt}, \binits{F.}},
\bauthor{\bsnm{Milani}, \binits{E.S.}},
\bauthor{\bsnm{Wollscheid}, \binits{B.}}:
\batitle{Elucidation of host-virus surfaceome interactions using spatial
  proteotyping.}
\bjtitle{Advances in Virus Research}
\bvolume{109},
\bfpage{105}--\blpage{134}
(\byear{2021})
\end{barticle}
\endbibitem

\bibitem{PMID:33934830}
\begin{barticle}
\bauthor{\bsnm{Zapatero-Belinch{\'o}n}, \binits{F.J.}},
\bauthor{\bsnm{Carriqu{\'\i}-Madro{\~n}al}, \binits{B.}},
\bauthor{\bsnm{Gerold}, \binits{G.}}:
\batitle{Proximity labeling approaches to study protein complexes during virus
  infection.}
\bjtitle{Advances in Virus Research}
\bvolume{109},
\bfpage{63}--\blpage{104}
(\byear{2021})
\end{barticle}
\endbibitem

\bibitem{PMID:29746851}
\begin{barticle}
\bauthor{\bsnm{Lasswitz}, \binits{L.}},
\bauthor{\bsnm{Chandra}, \binits{N.}},
\bauthor{\bsnm{Arnberg}, \binits{N.}},
\bauthor{\bsnm{Gerold}, \binits{G.}}:
\batitle{Glycomics and proteomics approaches to investigate early
  adenovirus--host cell interactions}.
\bjtitle{Journal of molecular biology}
\bvolume{430}(\bissue{13}),
\bfpage{1863}--\blpage{1882}
(\byear{2018})
\end{barticle}
\endbibitem

\bibitem{PMID:26365680}
\begin{barticle}
\bauthor{\bsnm{Gerold}, \binits{G.}},
\bauthor{\bsnm{Bruening}, \binits{J.}},
\bauthor{\bsnm{Pietschmann}, \binits{T.}}:
\batitle{Decoding protein networks during virus entry by quantitative
  proteomics}.
\bjtitle{Virus research}
\bvolume{218},
\bfpage{25}--\blpage{39}
(\byear{2016})
\end{barticle}
\endbibitem

\bibitem{PMID:26817613}
\begin{barticle}
\bauthor{\bsnm{Lum}, \binits{K.K.}},
\bauthor{\bsnm{Cristea}, \binits{I.M.}}:
\batitle{Proteomic approaches to uncovering virus--host protein interactions
  during the progression of viral infection}.
\bjtitle{Expert review of proteomics}
\bvolume{13}(\bissue{3}),
\bfpage{325}--\blpage{340}
(\byear{2016})
\end{barticle}
\endbibitem

\bibitem{PMID:28163258}
\begin{barticle}
\bauthor{\bsnm{Greco}, \binits{T.M.}},
\bauthor{\bsnm{Cristea}, \binits{I.M.}}:
\batitle{Proteomics tracing the footsteps of infectious disease}.
\bjtitle{Molecular \& Cellular Proteomics}
\bvolume{16}(\bissue{4}),
\bfpage{5}--\blpage{14}
(\byear{2017})
\end{barticle}
\endbibitem

\bibitem{PMID:28679763}
\begin{barticle}
\bauthor{\bsnm{Jean~Beltran}, \binits{P.M.}},
\bauthor{\bsnm{Cook}, \binits{K.C.}},
\bauthor{\bsnm{Cristea}, \binits{I.M.}}:
\batitle{Exploring and exploiting proteome organization during viral
  infection}.
\bjtitle{Journal of virology}
\bvolume{91}(\bissue{18}),
\bfpage{00268}--\blpage{17}
(\byear{2017})
\end{barticle}
\endbibitem

\bibitem{PMID:19632888}
\begin{barticle}
\bauthor{\bsnm{Bailer}, \binits{S.}},
\bauthor{\bsnm{Haas}, \binits{J.}}:
\batitle{Connecting viral with cellular interactomes}.
\bjtitle{Current opinion in microbiology}
\bvolume{12}(\bissue{4}),
\bfpage{453}--\blpage{459}
(\byear{2009})
\end{barticle}
\endbibitem

\bibitem{PMID:11967329}
\begin{barticle}
\bauthor{\bsnm{Spiropoulou}, \binits{C.F.}},
\bauthor{\bsnm{Kunz}, \binits{S.}},
\bauthor{\bsnm{Rollin}, \binits{P.E.}},
\bauthor{\bsnm{Campbell}, \binits{K.P.}},
\bauthor{\bsnm{Oldstone}, \binits{M.B.}}:
\batitle{New world arenavirus clade c, but not clade a and b viruses, utilizes
  $\alpha$-dystroglycan as its major receptor}.
\bjtitle{Journal of virology}
\bvolume{76}(\bissue{10}),
\bfpage{5140}--\blpage{5146}
(\byear{2002})
\end{barticle}
\endbibitem

\bibitem{kerrien2012intact}
\begin{barticle}
\bauthor{\bsnm{Kerrien}, \binits{S.}},
\bauthor{\bsnm{Aranda}, \binits{B.}},
\bauthor{\bsnm{Breuza}, \binits{L.}},
\bauthor{\bsnm{Bridge}, \binits{A.}},
\bauthor{\bsnm{Broackes-Carter}, \binits{F.}},
\bauthor{\bsnm{Chen}, \binits{C.}},
\bauthor{\bsnm{Duesbury}, \binits{M.}},
\bauthor{\bsnm{Dumousseau}, \binits{M.}},
\bauthor{\bsnm{Feuermann}, \binits{M.}},
\bauthor{\bsnm{Hinz}, \binits{U.}}, \betal:
\batitle{The intact molecular interaction database in 2012}.
\bjtitle{Nucleic acids research}
\bvolume{40}(\bissue{D1}),
\bfpage{841}--\blpage{846}
(\byear{2012})
\end{barticle}
\endbibitem

\bibitem{calderone2015virusmentha}
\begin{barticle}
\bauthor{\bsnm{Calderone}, \binits{A.}},
\bauthor{\bsnm{Licata}, \binits{L.}},
\bauthor{\bsnm{Cesareni}, \binits{G.}}:
\batitle{Virusmentha: a new resource for virus-host protein interactions}.
\bjtitle{Nucleic acids research}
\bvolume{43}(\bissue{D1}),
\bfpage{588}--\blpage{592}
(\byear{2015})
\end{barticle}
\endbibitem

\bibitem{chatr2009virusmint}
\begin{barticle}
\bauthor{\bsnm{Chatr-Aryamontri}, \binits{A.}},
\bauthor{\bsnm{Ceol}, \binits{A.}},
\bauthor{\bsnm{Peluso}, \binits{D.}},
\bauthor{\bsnm{Nardozza}, \binits{A.}},
\bauthor{\bsnm{Panni}, \binits{S.}},
\bauthor{\bsnm{Sacco}, \binits{F.}},
\bauthor{\bsnm{Tinti}, \binits{M.}},
\bauthor{\bsnm{Smolyar}, \binits{A.}},
\bauthor{\bsnm{Castagnoli}, \binits{L.}},
\bauthor{\bsnm{Vidal}, \binits{M.}}, \betal:
\batitle{Virusmint: a viral protein interaction database}.
\bjtitle{Nucleic acids research}
\bvolume{37}(\bissue{suppl\_1}),
\bfpage{669}--\blpage{673}
(\byear{2009})
\end{barticle}
\endbibitem

\bibitem{ammari2016hpidb}
\begin{botherref}
\oauthor{\bsnm{Ammari}, \binits{M.G.}},
\oauthor{\bsnm{Gresham}, \binits{C.R.}},
\oauthor{\bsnm{McCarthy}, \binits{F.M.}},
\oauthor{\bsnm{Nanduri}, \binits{B.}}:
Hpidb 2.0: a curated database for host--pathogen interactions.
Database
\textbf{2016}
(2016)
\end{botherref}
\endbibitem

\bibitem{requiao2020viruses}
\begin{barticle}
\bauthor{\bsnm{Requi{\~a}o}, \binits{R.D.}},
\bauthor{\bsnm{Carneiro}, \binits{R.L.}},
\bauthor{\bsnm{Moreira}, \binits{M.H.}},
\bauthor{\bsnm{Ribeiro-Alves}, \binits{M.}},
\bauthor{\bsnm{Rossetto}, \binits{S.}},
\bauthor{\bsnm{Palhano}, \binits{F.L.}},
\bauthor{\bsnm{Domitrovic}, \binits{T.}}:
\batitle{Viruses with different genome types adopt a similar strategy to pack
  nucleic acids based on positively charged protein domains}.
\bjtitle{Scientific reports}
\bvolume{10}(\bissue{1}),
\bfpage{1}--\blpage{12}
(\byear{2020})
\end{barticle}
\endbibitem

\bibitem{PMID:28007618}
\begin{barticle}
\bauthor{\bsnm{Rodrigo}, \binits{G.}},
\bauthor{\bsnm{Dar{\`o}s}, \binits{J.-A.}},
\bauthor{\bsnm{Elena}, \binits{S.F.}}:
\batitle{Virus-host interactome: putting the accent on how it changes}.
\bjtitle{Journal of proteomics}
\bvolume{156},
\bfpage{1}--\blpage{4}
(\byear{2017})
\end{barticle}
\endbibitem

\bibitem{PMID:25502394}
\begin{barticle}
\bauthor{\bsnm{Gitlin}, \binits{L.}},
\bauthor{\bsnm{Hagai}, \binits{T.}},
\bauthor{\bsnm{LaBarbera}, \binits{A.}},
\bauthor{\bsnm{Solovey}, \binits{M.}},
\bauthor{\bsnm{Andino}, \binits{R.}}:
\batitle{Rapid evolution of virus sequences in intrinsically disordered protein
  regions}.
\bjtitle{PLoS pathogens}
\bvolume{10}(\bissue{12}),
\bfpage{1004529}
(\byear{2014})
\end{barticle}
\endbibitem

\bibitem{eid2016denovo}
\begin{barticle}
\bauthor{\bsnm{Eid}, \binits{F.-E.}},
\bauthor{\bsnm{ElHefnawi}, \binits{M.}},
\bauthor{\bsnm{Heath}, \binits{L.S.}}:
\batitle{Denovo: virus-host sequence-based protein--protein interaction
  prediction}.
\bjtitle{Bioinformatics}
\bvolume{32}(\bissue{8}),
\bfpage{1144}--\blpage{1150}
(\byear{2016})
\end{barticle}
\endbibitem

\bibitem{li2020predicting}
\begin{botherref}
\oauthor{\bsnm{Li}, \binits{Y.}},
\oauthor{\bsnm{Ilie}, \binits{L.}}:
Predicting protein--protein interactions using sprint.
In: Protein-Protein Interaction Networks 2020,
pp. 1--11.
Springer
\end{botherref}
\endbibitem

\bibitem{sun2017sequence}
\begin{barticle}
\bauthor{\bsnm{Sun}, \binits{T.}},
\bauthor{\bsnm{Zhou}, \binits{B.}},
\bauthor{\bsnm{Lai}, \binits{L.}},
\bauthor{\bsnm{Pei}, \binits{J.}}:
\batitle{Sequence-based prediction of protein protein interaction using a
  deep-learning algorithm}.
\bjtitle{BMC bioinformatics}
\bvolume{18}(\bissue{1}),
\bfpage{1}--\blpage{8}
(\byear{2017})
\end{barticle}
\endbibitem

\bibitem{li2020computational}
\begin{botherref}
\oauthor{\bsnm{Li}, \binits{Y.}}:
Computational methods for predicting protein-protein interactions and binding
  sites
(2020)
\end{botherref}
\endbibitem

\bibitem{chen2019protein}
\begin{barticle}
\bauthor{\bsnm{Chen}, \binits{K.-H.}},
\bauthor{\bsnm{Wang}, \binits{T.-F.}},
\bauthor{\bsnm{Hu}, \binits{Y.-J.}}:
\batitle{Protein-protein interaction prediction using a hybrid feature
  representation and a stacked generalization scheme}.
\bjtitle{BMC bioinformatics}
\bvolume{20}(\bissue{1}),
\bfpage{1}--\blpage{17}
(\byear{2019})
\end{barticle}
\endbibitem

\bibitem{sarkar2019machine}
\begin{barticle}
\bauthor{\bsnm{Sarkar}, \binits{D.}},
\bauthor{\bsnm{Saha}, \binits{S.}}:
\batitle{Machine-learning techniques for the prediction of protein--protein
  interactions}.
\bjtitle{Journal of biosciences}
\bvolume{44}(\bissue{4}),
\bfpage{1}--\blpage{12}
(\byear{2019})
\end{barticle}
\endbibitem

\bibitem{sudhakar2020computational}
\begin{botherref}
\oauthor{\bsnm{Sudhakar}, \binits{P.}},
\oauthor{\bsnm{Machiels}, \binits{K.}},
\oauthor{\bsnm{Vermeire}, \binits{S.}}:
Computational biology and machine learning approaches to study mechanistic
  microbiomehost interactions
(2020)
\end{botherref}
\endbibitem

\bibitem{mei2020silico}
\begin{barticle}
\bauthor{\bsnm{Mei}, \binits{S.}},
\bauthor{\bsnm{Zhang}, \binits{K.}}:
\batitle{In silico unravelling pathogen-host signaling cross-talks via pathogen
  mimicry and human protein-protein interaction networks}.
\bjtitle{Computational and structural biotechnology journal}
\bvolume{18},
\bfpage{100}--\blpage{113}
(\byear{2020})
\end{barticle}
\endbibitem

\bibitem{dick2020pipe4}
\begin{barticle}
\bauthor{\bsnm{Dick}, \binits{K.}},
\bauthor{\bsnm{Samanfar}, \binits{B.}},
\bauthor{\bsnm{Barnes}, \binits{B.}},
\bauthor{\bsnm{Cober}, \binits{E.R.}},
\bauthor{\bsnm{Mimee}, \binits{B.}},
\bauthor{\bsnm{Molnar}, \binits{S.J.}},
\bauthor{\bsnm{Biggar}, \binits{K.K.}},
\bauthor{\bsnm{Golshani}, \binits{A.}},
\bauthor{\bsnm{Dehne}, \binits{F.}},
\bauthor{\bsnm{Green}, \binits{J.R.}}, \betal:
\batitle{Pipe4: Fast ppi predictor for comprehensive inter-and cross-species
  interactomes}.
\bjtitle{Scientific reports}
\bvolume{10}(\bissue{1}),
\bfpage{1}--\blpage{15}
(\byear{2020})
\end{barticle}
\endbibitem

\bibitem{li2014pathogen}
\begin{bchapter}
\bauthor{\bsnm{Li}, \binits{B.Y.S.}},
\bauthor{\bsnm{Yeung}, \binits{L.F.}},
\bauthor{\bsnm{Yang}, \binits{G.}}:
\bctitle{Pathogen host interaction prediction via matrix factorization}.
In: \bbtitle{2014 IEEE International Conference on Bioinformatics and
  Biomedicine (BIBM)},
pp. \bfpage{357}--\blpage{362}
(\byear{2014}).
\bcomment{IEEE}
\end{bchapter}
\endbibitem

\bibitem{guven2019interface}
\begin{botherref}
\oauthor{\bsnm{Guven-Maiorov}, \binits{E.}},
\oauthor{\bsnm{Tsai}, \binits{C.-J.}},
\oauthor{\bsnm{Ma}, \binits{B.}},
\oauthor{\bsnm{Nussinov}, \binits{R.}}:
Interface-based structural prediction of novel host-pathogen interactions.
In: Computational Methods in Protein Evolution 2019,
pp. 317--335.
Springer
\end{botherref}
\endbibitem

\bibitem{basit2018training}
\begin{barticle}
\bauthor{\bsnm{Basit}, \binits{A.H.}},
\bauthor{\bsnm{Abbasi}, \binits{W.A.}},
\bauthor{\bsnm{Asif}, \binits{A.}},
\bauthor{\bsnm{Gull}, \binits{S.}},
\bauthor{\bsnm{Minhas}, \binits{F.U.A.A.}}:
\batitle{Training host-pathogen protein--protein interaction predictors}.
\bjtitle{Journal of bioinformatics and computational biology}
\bvolume{16}(\bissue{04}),
\bfpage{1850014}
(\byear{2018})
\end{barticle}
\endbibitem

\bibitem{alley2019unified}
\begin{barticle}
\bauthor{\bsnm{Alley}, \binits{E.C.}},
\bauthor{\bsnm{Khimulya}, \binits{G.}},
\bauthor{\bsnm{Biswas}, \binits{S.}},
\bauthor{\bsnm{AlQuraishi}, \binits{M.}},
\bauthor{\bsnm{Church}, \binits{G.M.}}:
\batitle{Unified rational protein engineering with sequence-based deep
  representation learning}.
\bjtitle{Nature methods}
\bvolume{16}(\bissue{12}),
\bfpage{1315}--\blpage{1322}
(\byear{2019})
\end{barticle}
\endbibitem

\bibitem{nouretdinov2012determining}
\begin{botherref}
\oauthor{\bsnm{Nouretdinov}, \binits{I.}},
\oauthor{\bsnm{Gammerman}, \binits{A.}},
\oauthor{\bsnm{Qi}, \binits{Y.}},
\oauthor{\bsnm{Klein-Seetharaman}, \binits{J.}}:
Determining confidence of predicted interactions between hiv-1 and human
  proteins using conformal method.
In: Biocomputing 2012,
pp. 311--322.
World Scientific
\end{botherref}
\endbibitem

\bibitem{nourani2016computational}
\begin{barticle}
\bauthor{\bsnm{Nourani}, \binits{E.}},
\bauthor{\bsnm{Khunjush}, \binits{F.}},
\bauthor{\bsnm{Durmu{\c{s}}}, \binits{S.}}:
\batitle{Computational prediction of virus--human protein--protein interactions
  using embedding kernelized heterogeneous data}.
\bjtitle{Molecular BioSystems}
\bvolume{12}(\bissue{6}),
\bfpage{1976}--\blpage{1986}
(\byear{2016})
\end{barticle}
\endbibitem

\bibitem{mei2015novel}
\begin{barticle}
\bauthor{\bsnm{Mei}, \binits{S.}},
\bauthor{\bsnm{Zhu}, \binits{H.}}:
\batitle{A novel one-class svm based negative data sampling method for
  reconstructing proteome-wide htlv-human protein interaction networks}.
\bjtitle{Scientific reports}
\bvolume{5}(\bissue{1}),
\bfpage{1}--\blpage{13}
(\byear{2015})
\end{barticle}
\endbibitem

\bibitem{cui2012prediction}
\begin{bchapter}
\bauthor{\bsnm{Cui}, \binits{G.}},
\bauthor{\bsnm{Fang}, \binits{C.}},
\bauthor{\bsnm{Han}, \binits{K.}}:
\bctitle{Prediction of protein-protein interactions between viruses and human
  by an svm model}.
In: \bbtitle{BMC Bioinformatics},
vol. \bseriesno{13},
pp. \bfpage{1}--\blpage{10}
(\byear{2012}).
\bcomment{Springer}
\end{bchapter}
\endbibitem

\bibitem{kim2017improved}
\begin{barticle}
\bauthor{\bsnm{Kim}, \binits{B.}},
\bauthor{\bsnm{Alguwaizani}, \binits{S.}},
\bauthor{\bsnm{Zhou}, \binits{X.}},
\bauthor{\bsnm{Huang}, \binits{D.-S.}},
\bauthor{\bsnm{Park}, \binits{B.}},
\bauthor{\bsnm{Han}, \binits{K.}}:
\batitle{An improved method for predicting interactions between virus and human
  proteins}.
\bjtitle{Journal of bioinformatics and computational biology}
\bvolume{15}(\bissue{01}),
\bfpage{1650024}
(\byear{2017})
\end{barticle}
\endbibitem

\bibitem{loaiza2020predhpi}
\begin{botherref}
\oauthor{\bsnm{Loaiza}, \binits{C.D.}},
\oauthor{\bsnm{Kaundal}, \binits{R.}}:
Predhpi: an integrated web server platform for the detection and visualization
  of host--pathogen interactions using sequence-based methods.
Bioinformatics
(2020)
\end{botherref}
\endbibitem

\bibitem{zhou2018generalized}
\begin{barticle}
\bauthor{\bsnm{Zhou}, \binits{X.}},
\bauthor{\bsnm{Park}, \binits{B.}},
\bauthor{\bsnm{Choi}, \binits{D.}},
\bauthor{\bsnm{Han}, \binits{K.}}:
\batitle{A generalized approach to predicting protein-protein interactions
  between virus and host}.
\bjtitle{BMC genomics}
\bvolume{19}(\bissue{6}),
\bfpage{69}--\blpage{77}
(\byear{2018})
\end{barticle}
\endbibitem

\bibitem{ma2020seq}
\begin{botherref}
\oauthor{\bsnm{Ma}, \binits{Y.}},
\oauthor{\bsnm{He}, \binits{T.}},
\oauthor{\bsnm{Tan}, \binits{Y.-T.}}, et al.:
Seq-bel: Sequence-based ensemble learning for predicting virus-human
  protein-protein interaction.
IEEE/ACM Transactions on Computational Biology and Bioinformatics
(2020)
\end{botherref}
\endbibitem

\bibitem{deng2020predict}
\begin{bchapter}
\bauthor{\bsnm{Deng}, \binits{L.}},
\bauthor{\bsnm{Zhao}, \binits{J.}},
\bauthor{\bsnm{Zhang}, \binits{J.}}:
\bctitle{Predict the protein-protein interaction between virus and host through
  hybrid deep neural network}.
In: \bbtitle{2020 IEEE International Conference on Bioinformatics and
  Biomedicine (BIBM)},
pp. \bfpage{11}--\blpage{16}
(\byear{2020}).
\bcomment{IEEE}
\end{bchapter}
\endbibitem

\bibitem{dey2020machine}
\begin{barticle}
\bauthor{\bsnm{Dey}, \binits{L.}},
\bauthor{\bsnm{Chakraborty}, \binits{S.}},
\bauthor{\bsnm{Mukhopadhyay}, \binits{A.}}:
\batitle{Machine learning techniques for sequence-based prediction of
  viral--host interactions between sars-cov-2 and human proteins}.
\bjtitle{Biomedical journal}
\bvolume{43}(\bissue{5}),
\bfpage{438}--\blpage{450}
(\byear{2020})
\end{barticle}
\endbibitem

\bibitem{yang2020prediction}
\begin{barticle}
\bauthor{\bsnm{Yang}, \binits{X.}},
\bauthor{\bsnm{Yang}, \binits{S.}},
\bauthor{\bsnm{Li}, \binits{Q.}},
\bauthor{\bsnm{Wuchty}, \binits{S.}},
\bauthor{\bsnm{Zhang}, \binits{Z.}}:
\batitle{Prediction of human-virus protein-protein interactions through a
  sequence embedding-based machine learning method}.
\bjtitle{Computational and structural biotechnology journal}
\bvolume{18},
\bfpage{153}--\blpage{161}
(\byear{2020})
\end{barticle}
\endbibitem

\bibitem{lanchantin2020transfer}
\begin{botherref}
\oauthor{\bsnm{Lanchantin}, \binits{J.}},
\oauthor{\bsnm{Weingarten}, \binits{T.}},
\oauthor{\bsnm{Sekhon}, \binits{A.}},
\oauthor{\bsnm{Miller}, \binits{C.}},
\oauthor{\bsnm{Qi}, \binits{Y.}}:
Transfer learning for predicting virus-host protein interactions for novel
  virus sequences.
bioRxiv,
2020--12
(2021)
\end{botherref}
\endbibitem

\bibitem{wang2020prediction}
\begin{barticle}
\bauthor{\bsnm{Liu-Wei}, \binits{W.}},
\bauthor{\bsnm{Kafkas}, \binits{S.}},
\bauthor{\bsnm{Chen}, \binits{J.}},
\bauthor{\bsnm{Dimonaco}, \binits{N.J.}},
\bauthor{\bsnm{Tegnér}, \binits{J.}},
\bauthor{\bsnm{Hoehndorf}, \binits{R.}}:
\batitle{Deepviral: prediction of novel virus–host interactions from protein
  sequences and infectious disease phenotypes}.
\bjtitle{Bioinformatics}
(\byear{2021}).
doi:\doiurl{10.1093/bioinformatics/btab147}
\end{barticle}
\endbibitem

\bibitem{barman2014prediction}
\begin{barticle}
\bauthor{\bsnm{Barman}, \binits{R.K.}},
\bauthor{\bsnm{Saha}, \binits{S.}},
\bauthor{\bsnm{Das}, \binits{S.}}:
\batitle{Prediction of interactions between viral and host proteins using
  supervised machine learning methods}.
\bjtitle{PloS one}
\bvolume{9}(\bissue{11}),
\bfpage{112034}
(\byear{2014})
\end{barticle}
\endbibitem

\bibitem{lasso2019structure}
\begin{barticle}
\bauthor{\bsnm{Lasso}, \binits{G.}},
\bauthor{\bsnm{Mayer}, \binits{S.V.}},
\bauthor{\bsnm{Winkelmann}, \binits{E.R.}},
\bauthor{\bsnm{Chu}, \binits{T.}},
\bauthor{\bsnm{Elliot}, \binits{O.}},
\bauthor{\bsnm{Patino-Galindo}, \binits{J.A.}},
\bauthor{\bsnm{Park}, \binits{K.}},
\bauthor{\bsnm{Rabadan}, \binits{R.}},
\bauthor{\bsnm{Honig}, \binits{B.}},
\bauthor{\bsnm{Shapira}, \binits{S.D.}}:
\batitle{A structure-informed atlas of human-virus interactions}.
\bjtitle{Cell}
\bvolume{178}(\bissue{6}),
\bfpage{1526}--\blpage{1541}
(\byear{2019})
\end{barticle}
\endbibitem

\bibitem{liu2019predicting}
\begin{barticle}
\bauthor{\bsnm{Liu}, \binits{D.}},
\bauthor{\bsnm{Ma}, \binits{Y.}},
\bauthor{\bsnm{Jiang}, \binits{X.}},
\bauthor{\bsnm{He}, \binits{T.}}:
\batitle{Predicting virus-host association by kernelized logistic matrix
  factorization and similarity network fusion}.
\bjtitle{BMC bioinformatics}
\bvolume{20}(\bissue{16}),
\bfpage{1}--\blpage{10}
(\byear{2019})
\end{barticle}
\endbibitem

\bibitem{wang2020network}
\begin{barticle}
\bauthor{\bsnm{Wang}, \binits{W.}},
\bauthor{\bsnm{Ren}, \binits{J.}},
\bauthor{\bsnm{Tang}, \binits{K.}},
\bauthor{\bsnm{Dart}, \binits{E.}},
\bauthor{\bsnm{Ignacio-Espinoza}, \binits{J.C.}},
\bauthor{\bsnm{Fuhrman}, \binits{J.A.}},
\bauthor{\bsnm{Braun}, \binits{J.}},
\bauthor{\bsnm{Sun}, \binits{F.}},
\bauthor{\bsnm{Ahlgren}, \binits{N.A.}}:
\batitle{A network-based integrated framework for predicting virus--prokaryote
  interactions}.
\bjtitle{NAR genomics and bioinformatics}
\bvolume{2}(\bissue{2}),
\bfpage{044}
(\byear{2020})
\end{barticle}
\endbibitem

\bibitem{biswas2020principles}
\begin{botherref}
\oauthor{\bsnm{Biswas}, \binits{S.}}:
Principles of machine learning-guided protein engineering.
PhD thesis
(2020)
\end{botherref}
\endbibitem

\bibitem{szklarczyk2015string}
\begin{barticle}
\bauthor{\bsnm{Szklarczyk}, \binits{D.}},
\bauthor{\bsnm{Franceschini}, \binits{A.}},
\bauthor{\bsnm{Wyder}, \binits{S.}},
\bauthor{\bsnm{Forslund}, \binits{K.}},
\bauthor{\bsnm{Heller}, \binits{D.}},
\bauthor{\bsnm{Huerta-Cepas}, \binits{J.}},
\bauthor{\bsnm{Simonovic}, \binits{M.}},
\bauthor{\bsnm{Roth}, \binits{A.}},
\bauthor{\bsnm{Santos}, \binits{A.}},
\bauthor{\bsnm{Tsafou}, \binits{K.P.}}, \betal:
\batitle{String v10: protein--protein interaction networks, integrated over the
  tree of life}.
\bjtitle{Nucleic acids research}
\bvolume{43}(\bissue{D1}),
\bfpage{447}--\blpage{452}
(\byear{2015})
\end{barticle}
\endbibitem

\bibitem{alonso2016apid}
\begin{barticle}
\bauthor{\bsnm{Alonso-Lopez}, \binits{D.}},
\bauthor{\bsnm{Guti{\'e}rrez}, \binits{M.A.}},
\bauthor{\bsnm{Lopes}, \binits{K.P.}},
\bauthor{\bsnm{Prieto}, \binits{C.}},
\bauthor{\bsnm{Santamar{\'\i}a}, \binits{R.}},
\bauthor{\bsnm{De~Las~Rivas}, \binits{J.}}:
\batitle{Apid interactomes: providing proteome-based interactomes with
  controlled quality for multiple species and derived networks}.
\bjtitle{Nucleic acids research}
\bvolume{44}(\bissue{W1}),
\bfpage{529}--\blpage{535}
(\byear{2016})
\end{barticle}
\endbibitem

\bibitem{uniprot2015uniprot}
\begin{barticle}
\bauthor{\bsnm{Consortium}, \binits{U.}}:
\batitle{Uniprot: a hub for protein information}.
\bjtitle{Nucleic acids research}
\bvolume{43}(\bissue{D1}),
\bfpage{204}--\blpage{212}
(\byear{2015})
\end{barticle}
\endbibitem

\bibitem{aranda2011psicquic}
\begin{barticle}
\bauthor{\bsnm{Aranda}, \binits{B.}},
\bauthor{\bsnm{Blankenburg}, \binits{H.}},
\bauthor{\bsnm{Kerrien}, \binits{S.}},
\bauthor{\bsnm{Brinkman}, \binits{F.S.}},
\bauthor{\bsnm{Ceol}, \binits{A.}},
\bauthor{\bsnm{Chautard}, \binits{E.}},
\bauthor{\bsnm{Dana}, \binits{J.M.}},
\bauthor{\bsnm{De~Las~Rivas}, \binits{J.}},
\bauthor{\bsnm{Dumousseau}, \binits{M.}},
\bauthor{\bsnm{Galeota}, \binits{E.}}, \betal:
\batitle{Psicquic and psiscore: accessing and scoring molecular interactions}.
\bjtitle{Nature methods}
\bvolume{8}(\bissue{7}),
\bfpage{528}--\blpage{529}
(\byear{2011})
\end{barticle}
\endbibitem

\bibitem{martin2005predicting}
\begin{barticle}
\bauthor{\bsnm{Martin}, \binits{S.}},
\bauthor{\bsnm{Roe}, \binits{D.}},
\bauthor{\bsnm{Faulon}, \binits{J.-L.}}:
\batitle{Predicting protein--protein interactions using signature products}.
\bjtitle{Bioinformatics}
\bvolume{21}(\bissue{2}),
\bfpage{218}--\blpage{226}
(\byear{2005})
\end{barticle}
\endbibitem

\bibitem{mei2013probability}
\begin{barticle}
\bauthor{\bsnm{Mei}, \binits{S.}}:
\batitle{Probability weighted ensemble transfer learning for predicting
  interactions between hiv-1 and human proteins}.
\bjtitle{PLoS One}
\bvolume{8}(\bissue{11}),
\bfpage{79606}
(\byear{2013})
\end{barticle}
\endbibitem

\bibitem{federhen2012ncbi}
\begin{barticle}
\bauthor{\bsnm{Federhen}, \binits{S.}}:
\batitle{The ncbi taxonomy database}.
\bjtitle{Nucleic acids research}
\bvolume{40}(\bissue{D1}),
\bfpage{136}--\blpage{143}
(\byear{2012})
\end{barticle}
\endbibitem

\bibitem{diella2008understanding}
\begin{barticle}
\bauthor{\bsnm{Diella}, \binits{F.}},
\bauthor{\bsnm{Haslam}, \binits{N.}},
\bauthor{\bsnm{Chica}, \binits{C.}},
\bauthor{\bsnm{Budd}, \binits{A.}},
\bauthor{\bsnm{Michael}, \binits{S.}},
\bauthor{\bsnm{Brown}, \binits{N.P.}},
\bauthor{\bsnm{Trav{\'e}}, \binits{G.}},
\bauthor{\bsnm{Gibson}, \binits{T.J.}}:
\batitle{Understanding eukaryotic linear motifs and their role in cell
  signaling and regulation}.
\bjtitle{Front Biosci}
\bvolume{13}(\bissue{6580}),
\bfpage{603}
(\byear{2008})
\end{barticle}
\endbibitem

\bibitem{neduva2006peptides}
\begin{barticle}
\bauthor{\bsnm{Neduva}, \binits{V.}},
\bauthor{\bsnm{Russell}, \binits{R.B.}}:
\batitle{Peptides mediating interaction networks: new leads at last}.
\bjtitle{Current opinion in biotechnology}
\bvolume{17}(\bissue{5}),
\bfpage{465}--\blpage{471}
(\byear{2006})
\end{barticle}
\endbibitem

\bibitem{le2014distributed}
\begin{bchapter}
\bauthor{\bsnm{Le}, \binits{Q.}},
\bauthor{\bsnm{Mikolov}, \binits{T.}}:
\bctitle{Distributed representations of sentences and documents}.
In: \bbtitle{International Conference on Machine Learning},
pp. \bfpage{1188}--\blpage{1196}
(\byear{2014}).
\bcomment{PMLR}
\end{bchapter}
\endbibitem

\bibitem{pytorch}
\begin{barticle}
\bauthor{\bsnm{Paszke}, \binits{A.}},
\bauthor{\bsnm{Gross}, \binits{S.}},
\bauthor{\bsnm{Massa}, \binits{F.}},
\bauthor{\bsnm{Lerer}, \binits{A.}},
\bauthor{\bsnm{Bradbury}, \binits{J.}},
\bauthor{\bsnm{Chanan}, \binits{G.}},
\bauthor{\bsnm{Killeen}, \binits{T.}},
\bauthor{\bsnm{Lin}, \binits{Z.}},
\bauthor{\bsnm{Gimelshein}, \binits{N.}},
\bauthor{\bsnm{Antiga}, \binits{L.}}, \betal:
\batitle{Pytorch: An imperative style, high-performance deep learning library}.
\bjtitle{Advances in neural information processing systems}
\bvolume{32},
\bfpage{8026}--\blpage{8037}
(\byear{2019})
\end{barticle}
\endbibitem

\bibitem{welch1947generalization}
\begin{barticle}
\bauthor{\bsnm{Welch}, \binits{B.L.}}:
\batitle{The generalization of ‘student's’problem when several different
  population varlances are involved}.
\bjtitle{Biometrika}
\bvolume{34}(\bissue{1-2}),
\bfpage{28}--\blpage{35}
(\byear{1947})
\end{barticle}
\endbibitem

\bibitem{salzberg1997comparing}
\begin{barticle}
\bauthor{\bsnm{Salzberg}, \binits{S.L.}}:
\batitle{On comparing classifiers: Pitfalls to avoid and a recommended
  approach}.
\bjtitle{Data mining and knowledge discovery}
\bvolume{1}(\bissue{3}),
\bfpage{317}--\blpage{328}
(\byear{1997})
\end{barticle}
\endbibitem

\bibitem{kafadar1997handbook}
\begin{barticle}
\bauthor{\bsnm{Kafadar}, \binits{K.}}:
\batitle{Handbook of parametric and nonparametric statistical procedures}.
\bjtitle{The American Statistician}
\bvolume{51}(\bissue{4}),
\bfpage{374}
(\byear{1997})
\end{barticle}
\endbibitem

\bibitem{bausch2015mass}
\begin{barticle}
\bauthor{\bsnm{Bausch-Fluck}, \binits{D.}},
\bauthor{\bsnm{Hofmann}, \binits{A.}},
\bauthor{\bsnm{Bock}, \binits{T.}},
\bauthor{\bsnm{Frei}, \binits{A.P.}},
\bauthor{\bsnm{Cerciello}, \binits{F.}},
\bauthor{\bsnm{Jacobs}, \binits{A.}},
\bauthor{\bsnm{Moest}, \binits{H.}},
\bauthor{\bsnm{Omasits}, \binits{U.}},
\bauthor{\bsnm{Gundry}, \binits{R.L.}},
\bauthor{\bsnm{Yoon}, \binits{C.}}, \betal:
\batitle{A mass spectrometric-derived cell surface protein atlas}.
\bjtitle{PloS one}
\bvolume{10}(\bissue{4}),
\bfpage{0121314}
(\byear{2015})
\end{barticle}
\endbibitem

\bibitem{ashburner2000gene}
\begin{barticle}
\bauthor{\bsnm{Ashburner}, \binits{M.}},
\bauthor{\bsnm{Ball}, \binits{C.A.}},
\bauthor{\bsnm{Blake}, \binits{J.A.}},
\bauthor{\bsnm{Botstein}, \binits{D.}},
\bauthor{\bsnm{Butler}, \binits{H.}},
\bauthor{\bsnm{Cherry}, \binits{J.M.}},
\bauthor{\bsnm{Davis}, \binits{A.P.}},
\bauthor{\bsnm{Dolinski}, \binits{K.}},
\bauthor{\bsnm{Dwight}, \binits{S.S.}},
\bauthor{\bsnm{Eppig}, \binits{J.T.}}, \betal:
\batitle{Gene ontology: tool for the unification of biology}.
\bjtitle{Nature genetics}
\bvolume{25}(\bissue{1}),
\bfpage{25}--\blpage{29}
(\byear{2000})
\end{barticle}
\endbibitem

\bibitem{gene2021gene}
\begin{barticle}
\bauthor{\bsnm{Carbon}, \binits{S.}},
\bauthor{\bsnm{Douglass}, \binits{E.}},
\bauthor{\bsnm{Good}, \binits{B.M.}},
\bauthor{\bsnm{Unni}, \binits{D.R.}},
\bauthor{\bsnm{Harris}, \binits{N.L.}},
\bauthor{\bsnm{Mungall}, \binits{C.J.}},
\bauthor{\bsnm{Basu}, \binits{S.}},
\bauthor{\bsnm{Chisholm}, \binits{R.L.}},
\bauthor{\bsnm{Dodson}, \binits{R.J.}},
\bauthor{\bsnm{Hartline}, \binits{E.}}, \betal:
\batitle{The gene ontology resource: enriching a gold mine}.
\bjtitle{Nucleic Acids Research}
\bvolume{49}(\bissue{D1}),
\bfpage{325}--\blpage{334}
(\byear{2021})
\end{barticle}
\endbibitem

\bibitem{shang2020cell}
\begin{barticle}
\bauthor{\bsnm{Shang}, \binits{J.}},
\bauthor{\bsnm{Wan}, \binits{Y.}},
\bauthor{\bsnm{Luo}, \binits{C.}},
\bauthor{\bsnm{Ye}, \binits{G.}},
\bauthor{\bsnm{Geng}, \binits{Q.}},
\bauthor{\bsnm{Auerbach}, \binits{A.}},
\bauthor{\bsnm{Li}, \binits{F.}}:
\batitle{Cell entry mechanisms of sars-cov-2}.
\bjtitle{Proceedings of the National Academy of Sciences}
\bvolume{117}(\bissue{21}),
\bfpage{11727}--\blpage{11734}
(\byear{2020})
\end{barticle}
\endbibitem

\bibitem{zhang2021molecular}
\begin{barticle}
\bauthor{\bsnm{Zhang}, \binits{Q.}},
\bauthor{\bsnm{Xiang}, \binits{R.}},
\bauthor{\bsnm{Huo}, \binits{S.}},
\bauthor{\bsnm{Zhou}, \binits{Y.}},
\bauthor{\bsnm{Jiang}, \binits{S.}},
\bauthor{\bsnm{Wang}, \binits{Q.}},
\bauthor{\bsnm{Yu}, \binits{F.}}:
\batitle{Molecular mechanism of interaction between sars-cov-2 and host cells
  and interventional therapy}.
\bjtitle{Signal Transduction and Targeted Therapy}
\bvolume{6}(\bissue{1}),
\bfpage{1}--\blpage{19}
(\byear{2021})
\end{barticle}
\endbibitem

\bibitem{hoffmann2020sars}
\begin{barticle}
\bauthor{\bsnm{Hoffmann}, \binits{M.}},
\bauthor{\bsnm{Kleine-Weber}, \binits{H.}},
\bauthor{\bsnm{Schroeder}, \binits{S.}},
\bauthor{\bsnm{Kr{\"u}ger}, \binits{N.}},
\bauthor{\bsnm{Herrler}, \binits{T.}},
\bauthor{\bsnm{Erichsen}, \binits{S.}},
\bauthor{\bsnm{Schiergens}, \binits{T.S.}},
\bauthor{\bsnm{Herrler}, \binits{G.}},
\bauthor{\bsnm{Wu}, \binits{N.-H.}},
\bauthor{\bsnm{Nitsche}, \binits{A.}}, \betal:
\batitle{Sars-cov-2 cell entry depends on ace2 and tmprss2 and is blocked by a
  clinically proven protease inhibitor}.
\bjtitle{cell}
\bvolume{181}(\bissue{2}),
\bfpage{271}--\blpage{280}
(\byear{2020})
\end{barticle}
\endbibitem

\bibitem{PMID:14647384}
\begin{barticle}
\bauthor{\bsnm{Li}, \binits{W.}},
\bauthor{\bsnm{Moore}, \binits{M.J.}},
\bauthor{\bsnm{Vasilieva}, \binits{N.}},
\bauthor{\bsnm{Sui}, \binits{J.}},
\bauthor{\bsnm{Wong}, \binits{S.K.}},
\bauthor{\bsnm{Berne}, \binits{M.A.}},
\bauthor{\bsnm{Somasundaran}, \binits{M.}},
\bauthor{\bsnm{Sullivan}, \binits{J.L.}},
\bauthor{\bsnm{Luzuriaga}, \binits{K.}},
\bauthor{\bsnm{Greenough}, \binits{T.C.}}, \betal:
\batitle{Angiotensin-converting enzyme 2 is a functional receptor for the sars
  coronavirus}.
\bjtitle{Nature}
\bvolume{426}(\bissue{6965}),
\bfpage{450}--\blpage{454}
(\byear{2003})
\end{barticle}
\endbibitem

\bibitem{PMID:32380511}
\begin{barticle}
\bauthor{\bsnm{Bao}, \binits{L.}},
\bauthor{\bsnm{Deng}, \binits{W.}},
\bauthor{\bsnm{Huang}, \binits{B.}},
\bauthor{\bsnm{Gao}, \binits{H.}},
\bauthor{\bsnm{Liu}, \binits{J.}},
\bauthor{\bsnm{Ren}, \binits{L.}},
\bauthor{\bsnm{Wei}, \binits{Q.}},
\bauthor{\bsnm{Yu}, \binits{P.}},
\bauthor{\bsnm{Xu}, \binits{Y.}},
\bauthor{\bsnm{Qi}, \binits{F.}}, \betal:
\batitle{The pathogenicity of sars-cov-2 in hace2 transgenic mice}.
\bjtitle{Nature}
\bvolume{583}(\bissue{7818}),
\bfpage{830}--\blpage{833}
(\byear{2020})
\end{barticle}
\endbibitem

\bibitem{PMID:32839612}
\begin{barticle}
\bauthor{\bsnm{Winkler}, \binits{E.S.}},
\bauthor{\bsnm{Bailey}, \binits{A.L.}},
\bauthor{\bsnm{Kafai}, \binits{N.M.}},
\bauthor{\bsnm{Nair}, \binits{S.}},
\bauthor{\bsnm{McCune}, \binits{B.T.}},
\bauthor{\bsnm{Yu}, \binits{J.}},
\bauthor{\bsnm{Fox}, \binits{J.M.}},
\bauthor{\bsnm{Chen}, \binits{R.E.}},
\bauthor{\bsnm{Earnest}, \binits{J.T.}},
\bauthor{\bsnm{Keeler}, \binits{S.P.}}, \betal:
\batitle{Sars-cov-2 infection of human ace2-transgenic mice causes severe lung
  inflammation and impaired function}.
\bjtitle{Nature immunology}
\bvolume{21}(\bissue{11}),
\bfpage{1327}--\blpage{1335}
(\byear{2020})
\end{barticle}
\endbibitem

\bibitem{wang2013structure}
\begin{barticle}
\bauthor{\bsnm{Wang}, \binits{N.}},
\bauthor{\bsnm{Shi}, \binits{X.}},
\bauthor{\bsnm{Jiang}, \binits{L.}},
\bauthor{\bsnm{Zhang}, \binits{S.}},
\bauthor{\bsnm{Wang}, \binits{D.}},
\bauthor{\bsnm{Tong}, \binits{P.}},
\bauthor{\bsnm{Guo}, \binits{D.}},
\bauthor{\bsnm{Fu}, \binits{L.}},
\bauthor{\bsnm{Cui}, \binits{Y.}},
\bauthor{\bsnm{Liu}, \binits{X.}}, \betal:
\batitle{Structure of mers-cov spike receptor-binding domain complexed with
  human receptor dpp4}.
\bjtitle{Cell research}
\bvolume{23}(\bissue{8}),
\bfpage{986}--\blpage{993}
(\byear{2013})
\end{barticle}
\endbibitem

\bibitem{vankadari2020emerging}
\begin{barticle}
\bauthor{\bsnm{Vankadari}, \binits{N.}},
\bauthor{\bsnm{Wilce}, \binits{J.A.}}:
\batitle{Emerging covid-19 coronavirus: glycan shield and structure prediction
  of spike glycoprotein and its interaction with human cd26}.
\bjtitle{Emerging microbes \& infections}
\bvolume{9}(\bissue{1}),
\bfpage{601}--\blpage{604}
(\byear{2020})
\end{barticle}
\endbibitem

\bibitem{PMID:1350662}
\begin{barticle}
\bauthor{\bsnm{Yeager}, \binits{C.L.}},
\bauthor{\bsnm{Ashmun}, \binits{R.A.}},
\bauthor{\bsnm{Williams}, \binits{R.K.}},
\bauthor{\bsnm{Cardellichio}, \binits{C.B.}},
\bauthor{\bsnm{Shapiro}, \binits{L.H.}},
\bauthor{\bsnm{Look}, \binits{A.T.}},
\bauthor{\bsnm{Holmes}, \binits{K.V.}}:
\batitle{Human aminopeptidase n is a receptor for human coronavirus 229e}.
\bjtitle{Nature}
\bvolume{357}(\bissue{6377}),
\bfpage{420}--\blpage{422}
(\byear{1992})
\end{barticle}
\endbibitem

\bibitem{PMID:34265844}
\begin{botherref}
\oauthor{\bsnm{Jumper}, \binits{J.}},
\oauthor{\bsnm{Evans}, \binits{R.}},
\oauthor{\bsnm{Pritzel}, \binits{A.}},
\oauthor{\bsnm{Green}, \binits{T.}},
\oauthor{\bsnm{Figurnov}, \binits{M.}},
\oauthor{\bsnm{Ronneberger}, \binits{O.}},
\oauthor{\bsnm{Tunyasuvunakool}, \binits{K.}},
\oauthor{\bsnm{Bates}, \binits{R.}},
\oauthor{\bsnm{{\v{Z}}{\'\i}dek}, \binits{A.}},
\oauthor{\bsnm{Potapenko}, \binits{A.}}, et al.:
Highly accurate protein structure prediction with alphafold.
Nature,
1--11
(2021)
\end{botherref}
\endbibitem

\bibitem{dong2021multitask}
\begin{barticle}
\bauthor{\bsnm{Dong}, \binits{N.T.}},
\bauthor{\bsnm{Khosla}, \binits{M.}}:
\batitle{A multitask transfer learning framework for novel virus-human protein
  interactions}.
\bjtitle{bioRxiv}
(\byear{2021}).
doi:\doiurl{10.1101/2021.03.25.437037}
\end{barticle}
\endbibitem

\end{thebibliography}

\newcommand{\BMCxmlcomment}[1]{}

\BMCxmlcomment{

<refgrp>

<bibl id="B1">
  <title><p>Comparing SARS-CoV-2 with SARS-CoV and influenza
  pandemics</p></title>
  <aug>
    <au><snm>Petersen</snm><fnm>E</fnm></au>
    <au><snm>Koopmans</snm><fnm>M</fnm></au>
    <au><snm>Go</snm><fnm>U</fnm></au>
    <au><snm>Hamer</snm><fnm>HH</fnm></au>
    <au><snm>Petrosillo</snm><fnm>N</fnm></au>
    <au><snm>Castelli</snm><fnm>F</fnm></au>
    <au><snm>Storgaard</snm><fnm>M</fnm></au>
    <au><snm>Al Khalili</snm><fnm>S</fnm></au>
    <au><snm>Simonsen</snm><fnm>L</fnm></au>
  </aug>
  <source>The Lancet Infectious Diseases</source>
  <publisher>Elsevier</publisher>
  <pubdate>2020</pubdate>
  <volume>20</volume>
  <issue>9</issue>
  <fpage>e238</fpage>
  <lpage>2244</lpage>
</bibl>

<bibl id="B2">
  <title><p>Break ins and break outs: viral interactions with the cytoskeleton
  of Mammalian cells</p></title>
  <aug>
    <au><snm>Smith</snm><fnm>GA</fnm></au>
    <au><snm>Enquist</snm><fnm>LW</fnm></au>
  </aug>
  <source>Annual Review of cell and developmental biology</source>
  <publisher>Annual Reviews</publisher>
  <pubdate>2002</pubdate>
  <volume>18</volume>
  <fpage>135</fpage>
  <lpage>61</lpage>
</bibl>

<bibl id="B3">
  <title><p>Exploring and Exploiting Proteome Organization during Viral
  Infection</p></title>
  <aug>
    <au><snm>Beltran</snm><fnm>PMJ</fnm></au>
    <au><snm>Cook</snm><fnm>KC</fnm></au>
    <au><snm>Cristea</snm><fnm>IM</fnm></au>
  </aug>
  <source>Journal of Virology</source>
  <publisher>American Society for Microbiology</publisher>
  <pubdate>2017</pubdate>
  <volume>91</volume>
  <issue>18</issue>
  <fpage>e00268</fpage>
  <lpage>17</lpage>
</bibl>

<bibl id="B4">
  <title><p>Protein Interactions during the Flavivirus and Hepacivirus Life
  Cycle</p></title>
  <aug>
    <au><snm>Gerold</snm><fnm>G</fnm></au>
    <au><snm>Bruening</snm><fnm>J</fnm></au>
    <au><snm>Weigel</snm><fnm>B</fnm></au>
    <au><snm>Pietschmann</snm><fnm>T</fnm></au>
  </aug>
  <source>Molecular and Cellular Proteomics</source>
  <publisher>American Society for Biochemistry and Molecular
  Biology</publisher>
  <pubdate>2017</pubdate>
  <volume>16</volume>
  <issue>4 suppl 1</issue>
  <fpage>S75</fpage>
  <lpage>S91</lpage>
</bibl>

<bibl id="B5">
  <title><p>Exploring the SARS-CoV-2 virus-host-drug interactome for drug
  repurposing</p></title>
  <aug>
    <au><snm>Sadegh</snm><fnm>S</fnm></au>
    <au><snm>Matschinske</snm><fnm>J</fnm></au>
    <au><snm>Blumenthal</snm><fnm>DB</fnm></au>
    <au><snm>Galindez</snm><fnm>G</fnm></au>
    <au><snm>Kacprowski</snm><fnm>T</fnm></au>
    <au><snm>List</snm><fnm>M</fnm></au>
    <au><snm>Nasirigerdeh</snm><fnm>R</fnm></au>
    <au><snm>Oubounyt</snm><fnm>M</fnm></au>
    <au><snm>Pichlmair</snm><fnm>A</fnm></au>
    <au><snm>Rose</snm><fnm>TD</fnm></au>
    <au><cnm>others</cnm></au>
  </aug>
  <source>Nature communications</source>
  <publisher>Nature Publishing Group</publisher>
  <pubdate>2020</pubdate>
  <volume>11</volume>
  <issue>1</issue>
  <fpage>1</fpage>
  <lpage>-9</lpage>
</bibl>

<bibl id="B6">
  <title><p>Elucidation of host-virus surfaceome interactions using spatial
  proteotyping.</p></title>
  <aug>
    <au><snm>Wendt</snm><fnm>F</fnm></au>
    <au><snm>Milani</snm><fnm>ES</fnm></au>
    <au><snm>Wollscheid</snm><fnm>B</fnm></au>
  </aug>
  <source>Advances in Virus Research</source>
  <pubdate>2021</pubdate>
  <volume>109</volume>
  <fpage>105</fpage>
  <lpage>-134</lpage>
</bibl>

<bibl id="B7">
  <title><p>Proximity labeling approaches to study protein complexes during
  virus infection.</p></title>
  <aug>
    <au><snm>Zapatero Belinch{\'o}n</snm><fnm>FJ</fnm></au>
    <au><snm>Carriqu{\'\i} Madro{\~n}al</snm><fnm>B</fnm></au>
    <au><snm>Gerold</snm><fnm>G</fnm></au>
  </aug>
  <source>Advances in Virus Research</source>
  <pubdate>2021</pubdate>
  <volume>109</volume>
  <fpage>63</fpage>
  <lpage>-104</lpage>
</bibl>

<bibl id="B8">
  <title><p>Glycomics and proteomics approaches to investigate early
  adenovirus--host cell interactions</p></title>
  <aug>
    <au><snm>Lasswitz</snm><fnm>L</fnm></au>
    <au><snm>Chandra</snm><fnm>N</fnm></au>
    <au><snm>Arnberg</snm><fnm>N</fnm></au>
    <au><snm>Gerold</snm><fnm>G</fnm></au>
  </aug>
  <source>Journal of molecular biology</source>
  <publisher>Elsevier</publisher>
  <pubdate>2018</pubdate>
  <volume>430</volume>
  <issue>13</issue>
  <fpage>1863</fpage>
  <lpage>-1882</lpage>
</bibl>

<bibl id="B9">
  <title><p>Decoding protein networks during virus entry by quantitative
  proteomics</p></title>
  <aug>
    <au><snm>Gerold</snm><fnm>G</fnm></au>
    <au><snm>Bruening</snm><fnm>J</fnm></au>
    <au><snm>Pietschmann</snm><fnm>T</fnm></au>
  </aug>
  <source>Virus research</source>
  <publisher>Elsevier</publisher>
  <pubdate>2016</pubdate>
  <volume>218</volume>
  <fpage>25</fpage>
  <lpage>-39</lpage>
</bibl>

<bibl id="B10">
  <title><p>Proteomic approaches to uncovering virus--host protein interactions
  during the progression of viral infection</p></title>
  <aug>
    <au><snm>Lum</snm><fnm>KK</fnm></au>
    <au><snm>Cristea</snm><fnm>IM</fnm></au>
  </aug>
  <source>Expert review of proteomics</source>
  <publisher>Taylor \& Francis</publisher>
  <pubdate>2016</pubdate>
  <volume>13</volume>
  <issue>3</issue>
  <fpage>325</fpage>
  <lpage>-340</lpage>
</bibl>

<bibl id="B11">
  <title><p>Proteomics tracing the footsteps of infectious disease</p></title>
  <aug>
    <au><snm>Greco</snm><fnm>TM</fnm></au>
    <au><snm>Cristea</snm><fnm>IM</fnm></au>
  </aug>
  <source>Molecular \& Cellular Proteomics</source>
  <publisher>ASBMB</publisher>
  <pubdate>2017</pubdate>
  <volume>16</volume>
  <issue>4</issue>
  <fpage>S5</fpage>
  <lpage>-S14</lpage>
</bibl>

<bibl id="B12">
  <title><p>Exploring and exploiting proteome organization during viral
  infection</p></title>
  <aug>
    <au><snm>Jean Beltran</snm><fnm>PM</fnm></au>
    <au><snm>Cook</snm><fnm>KC</fnm></au>
    <au><snm>Cristea</snm><fnm>IM</fnm></au>
  </aug>
  <source>Journal of virology</source>
  <publisher>Am Soc Microbiol</publisher>
  <pubdate>2017</pubdate>
  <volume>91</volume>
  <issue>18</issue>
  <fpage>e00268</fpage>
  <lpage>-17</lpage>
</bibl>

<bibl id="B13">
  <title><p>Connecting viral with cellular interactomes</p></title>
  <aug>
    <au><snm>Bailer</snm><fnm>SM</fnm></au>
    <au><snm>Haas</snm><fnm>J</fnm></au>
  </aug>
  <source>Current opinion in microbiology</source>
  <publisher>Elsevier</publisher>
  <pubdate>2009</pubdate>
  <volume>12</volume>
  <issue>4</issue>
  <fpage>453</fpage>
  <lpage>-459</lpage>
</bibl>

<bibl id="B14">
  <title><p>New World arenavirus clade C, but not clade A and B viruses,
  utilizes $\alpha$-dystroglycan as its major receptor</p></title>
  <aug>
    <au><snm>Spiropoulou</snm><fnm>CF</fnm></au>
    <au><snm>Kunz</snm><fnm>S</fnm></au>
    <au><snm>Rollin</snm><fnm>PE</fnm></au>
    <au><snm>Campbell</snm><fnm>KP</fnm></au>
    <au><snm>Oldstone</snm><fnm>MB</fnm></au>
  </aug>
  <source>Journal of virology</source>
  <publisher>Am Soc Microbiol</publisher>
  <pubdate>2002</pubdate>
  <volume>76</volume>
  <issue>10</issue>
  <fpage>5140</fpage>
  <lpage>-5146</lpage>
</bibl>

<bibl id="B15">
  <title><p>The IntAct molecular interaction database in 2012</p></title>
  <aug>
    <au><snm>Kerrien</snm><fnm>S</fnm></au>
    <au><snm>Aranda</snm><fnm>B</fnm></au>
    <au><snm>Breuza</snm><fnm>L</fnm></au>
    <au><snm>Bridge</snm><fnm>A</fnm></au>
    <au><snm>Broackes Carter</snm><fnm>F</fnm></au>
    <au><snm>Chen</snm><fnm>C</fnm></au>
    <au><snm>Duesbury</snm><fnm>M</fnm></au>
    <au><snm>Dumousseau</snm><fnm>M</fnm></au>
    <au><snm>Feuermann</snm><fnm>M</fnm></au>
    <au><snm>Hinz</snm><fnm>U</fnm></au>
    <au><cnm>others</cnm></au>
  </aug>
  <source>Nucleic acids research</source>
  <publisher>Oxford University Press</publisher>
  <pubdate>2012</pubdate>
  <volume>40</volume>
  <issue>D1</issue>
  <fpage>D841</fpage>
  <lpage>-D846</lpage>
</bibl>

<bibl id="B16">
  <title><p>VirusMentha: a new resource for virus-host protein
  interactions</p></title>
  <aug>
    <au><snm>Calderone</snm><fnm>A</fnm></au>
    <au><snm>Licata</snm><fnm>L</fnm></au>
    <au><snm>Cesareni</snm><fnm>G</fnm></au>
  </aug>
  <source>Nucleic acids research</source>
  <publisher>Oxford University Press</publisher>
  <pubdate>2015</pubdate>
  <volume>43</volume>
  <issue>D1</issue>
  <fpage>D588</fpage>
  <lpage>-D592</lpage>
</bibl>

<bibl id="B17">
  <title><p>VirusMINT: a viral protein interaction database</p></title>
  <aug>
    <au><snm>Chatr Aryamontri</snm><fnm>A</fnm></au>
    <au><snm>Ceol</snm><fnm>A</fnm></au>
    <au><snm>Peluso</snm><fnm>D</fnm></au>
    <au><snm>Nardozza</snm><fnm>A</fnm></au>
    <au><snm>Panni</snm><fnm>S</fnm></au>
    <au><snm>Sacco</snm><fnm>F</fnm></au>
    <au><snm>Tinti</snm><fnm>M</fnm></au>
    <au><snm>Smolyar</snm><fnm>A</fnm></au>
    <au><snm>Castagnoli</snm><fnm>L</fnm></au>
    <au><snm>Vidal</snm><fnm>M</fnm></au>
    <au><cnm>others</cnm></au>
  </aug>
  <source>Nucleic acids research</source>
  <publisher>Oxford University Press</publisher>
  <pubdate>2009</pubdate>
  <volume>37</volume>
  <issue>suppl\_1</issue>
  <fpage>D669</fpage>
  <lpage>-D673</lpage>
</bibl>

<bibl id="B18">
  <title><p>HPIDB 2.0: a curated database for host--pathogen
  interactions</p></title>
  <aug>
    <au><snm>Ammari</snm><fnm>MG</fnm></au>
    <au><snm>Gresham</snm><fnm>CR</fnm></au>
    <au><snm>McCarthy</snm><fnm>FM</fnm></au>
    <au><snm>Nanduri</snm><fnm>B</fnm></au>
  </aug>
  <source>Database</source>
  <publisher>Oxford Academic</publisher>
  <pubdate>2016</pubdate>
  <volume>2016</volume>
</bibl>

<bibl id="B19">
  <title><p>Viruses with different genome types adopt a similar strategy to
  pack nucleic acids based on positively charged protein domains</p></title>
  <aug>
    <au><snm>Requi{\~a}o</snm><fnm>RD</fnm></au>
    <au><snm>Carneiro</snm><fnm>RL</fnm></au>
    <au><snm>Moreira</snm><fnm>MH</fnm></au>
    <au><snm>Ribeiro Alves</snm><fnm>M</fnm></au>
    <au><snm>Rossetto</snm><fnm>S</fnm></au>
    <au><snm>Palhano</snm><fnm>FL</fnm></au>
    <au><snm>Domitrovic</snm><fnm>T</fnm></au>
  </aug>
  <source>Scientific reports</source>
  <publisher>Nature Publishing Group</publisher>
  <pubdate>2020</pubdate>
  <volume>10</volume>
  <issue>1</issue>
  <fpage>1</fpage>
  <lpage>-12</lpage>
</bibl>

<bibl id="B20">
  <title><p>Virus-host interactome: putting the accent on how it
  changes</p></title>
  <aug>
    <au><snm>Rodrigo</snm><fnm>G</fnm></au>
    <au><snm>Dar{\`o}s</snm><fnm>JA</fnm></au>
    <au><snm>Elena</snm><fnm>SF</fnm></au>
  </aug>
  <source>Journal of proteomics</source>
  <publisher>Elsevier</publisher>
  <pubdate>2017</pubdate>
  <volume>156</volume>
  <fpage>1</fpage>
  <lpage>-4</lpage>
</bibl>

<bibl id="B21">
  <title><p>Rapid evolution of virus sequences in intrinsically disordered
  protein regions</p></title>
  <aug>
    <au><snm>Gitlin</snm><fnm>L</fnm></au>
    <au><snm>Hagai</snm><fnm>T</fnm></au>
    <au><snm>LaBarbera</snm><fnm>A</fnm></au>
    <au><snm>Solovey</snm><fnm>M</fnm></au>
    <au><snm>Andino</snm><fnm>R</fnm></au>
  </aug>
  <source>PLoS pathogens</source>
  <publisher>Public Library of Science San Francisco, USA</publisher>
  <pubdate>2014</pubdate>
  <volume>10</volume>
  <issue>12</issue>
  <fpage>e1004529</fpage>
</bibl>

<bibl id="B22">
  <title><p>DeNovo: virus-host sequence-based protein--protein interaction
  prediction</p></title>
  <aug>
    <au><snm>Eid</snm><fnm>FE</fnm></au>
    <au><snm>ElHefnawi</snm><fnm>M</fnm></au>
    <au><snm>Heath</snm><fnm>LS</fnm></au>
  </aug>
  <source>Bioinformatics</source>
  <publisher>Oxford University Press</publisher>
  <pubdate>2016</pubdate>
  <volume>32</volume>
  <issue>8</issue>
  <fpage>1144</fpage>
  <lpage>-1150</lpage>
</bibl>

<bibl id="B23">
  <title><p>Predicting protein--protein interactions using sprint</p></title>
  <aug>
    <au><snm>Li</snm><fnm>Y</fnm></au>
    <au><snm>Ilie</snm><fnm>L</fnm></au>
  </aug>
  <source>Protein-Protein Interaction Networks 2020</source>
  <publisher>Springer</publisher>
  <fpage>1</fpage>
  <lpage>-11</lpage>
</bibl>

<bibl id="B24">
  <title><p>Sequence-based prediction of protein protein interaction using a
  deep-learning algorithm</p></title>
  <aug>
    <au><snm>Sun</snm><fnm>T</fnm></au>
    <au><snm>Zhou</snm><fnm>B</fnm></au>
    <au><snm>Lai</snm><fnm>L</fnm></au>
    <au><snm>Pei</snm><fnm>J</fnm></au>
  </aug>
  <source>BMC bioinformatics</source>
  <publisher>BioMed Central</publisher>
  <pubdate>2017</pubdate>
  <volume>18</volume>
  <issue>1</issue>
  <fpage>1</fpage>
  <lpage>-8</lpage>
</bibl>

<bibl id="B25">
  <title><p>Computational Methods for Predicting Protein-protein Interactions
  and Binding Sites</p></title>
  <aug>
    <au><snm>Li</snm><fnm>Y</fnm></au>
  </aug>
  <pubdate>2020</pubdate>
</bibl>

<bibl id="B26">
  <title><p>Protein-protein interaction prediction using a hybrid feature
  representation and a stacked generalization scheme</p></title>
  <aug>
    <au><snm>Chen</snm><fnm>KH</fnm></au>
    <au><snm>Wang</snm><fnm>TF</fnm></au>
    <au><snm>Hu</snm><fnm>YJ</fnm></au>
  </aug>
  <source>BMC bioinformatics</source>
  <publisher>Springer</publisher>
  <pubdate>2019</pubdate>
  <volume>20</volume>
  <issue>1</issue>
  <fpage>1</fpage>
  <lpage>-17</lpage>
</bibl>

<bibl id="B27">
  <title><p>Machine-learning techniques for the prediction of protein--protein
  interactions</p></title>
  <aug>
    <au><snm>Sarkar</snm><fnm>D</fnm></au>
    <au><snm>Saha</snm><fnm>S</fnm></au>
  </aug>
  <source>Journal of biosciences</source>
  <publisher>Springer</publisher>
  <pubdate>2019</pubdate>
  <volume>44</volume>
  <issue>4</issue>
  <fpage>1</fpage>
  <lpage>-12</lpage>
</bibl>

<bibl id="B28">
  <title><p>Computational Biology and Machine Learning Approaches to Study
  Mechanistic Microbiomehost Interactions</p></title>
  <aug>
    <au><snm>Sudhakar</snm><fnm>P</fnm></au>
    <au><snm>Machiels</snm><fnm>K</fnm></au>
    <au><snm>Vermeire</snm><fnm>S</fnm></au>
  </aug>
  <publisher>Preprints</publisher>
  <pubdate>2020</pubdate>
</bibl>

<bibl id="B29">
  <title><p>In silico unravelling pathogen-host signaling cross-talks via
  pathogen mimicry and human protein-protein interaction networks</p></title>
  <aug>
    <au><snm>Mei</snm><fnm>S</fnm></au>
    <au><snm>Zhang</snm><fnm>K</fnm></au>
  </aug>
  <source>Computational and structural biotechnology journal</source>
  <publisher>Elsevier</publisher>
  <pubdate>2020</pubdate>
  <volume>18</volume>
  <fpage>100</fpage>
  <lpage>-113</lpage>
</bibl>

<bibl id="B30">
  <title><p>Pipe4: Fast ppi predictor for comprehensive inter-and cross-species
  interactomes</p></title>
  <aug>
    <au><snm>Dick</snm><fnm>K</fnm></au>
    <au><snm>Samanfar</snm><fnm>B</fnm></au>
    <au><snm>Barnes</snm><fnm>B</fnm></au>
    <au><snm>Cober</snm><fnm>ER</fnm></au>
    <au><snm>Mimee</snm><fnm>B</fnm></au>
    <au><snm>Molnar</snm><fnm>SJ</fnm></au>
    <au><snm>Biggar</snm><fnm>KK</fnm></au>
    <au><snm>Golshani</snm><fnm>A</fnm></au>
    <au><snm>Dehne</snm><fnm>F</fnm></au>
    <au><snm>Green</snm><fnm>JR</fnm></au>
    <au><cnm>others</cnm></au>
  </aug>
  <source>Scientific reports</source>
  <publisher>Nature Publishing Group</publisher>
  <pubdate>2020</pubdate>
  <volume>10</volume>
  <issue>1</issue>
  <fpage>1</fpage>
  <lpage>-15</lpage>
</bibl>

<bibl id="B31">
  <title><p>Pathogen host interaction prediction via matrix
  factorization</p></title>
  <aug>
    <au><snm>Li</snm><fnm>BYS</fnm></au>
    <au><snm>Yeung</snm><fnm>LF</fnm></au>
    <au><snm>Yang</snm><fnm>G</fnm></au>
  </aug>
  <source>2014 IEEE International Conference on Bioinformatics and Biomedicine
  (BIBM)</source>
  <pubdate>2014</pubdate>
  <fpage>357</fpage>
  <lpage>-362</lpage>
</bibl>

<bibl id="B32">
  <title><p>Interface-based structural prediction of novel host-pathogen
  interactions</p></title>
  <aug>
    <au><snm>Guven Maiorov</snm><fnm>E</fnm></au>
    <au><snm>Tsai</snm><fnm>CJ</fnm></au>
    <au><snm>Ma</snm><fnm>B</fnm></au>
    <au><snm>Nussinov</snm><fnm>R</fnm></au>
  </aug>
  <source>Computational Methods in Protein Evolution 2019</source>
  <publisher>Springer</publisher>
  <fpage>317</fpage>
  <lpage>-335</lpage>
</bibl>

<bibl id="B33">
  <title><p>Training host-pathogen protein--protein interaction
  predictors</p></title>
  <aug>
    <au><snm>Basit</snm><fnm>AH</fnm></au>
    <au><snm>Abbasi</snm><fnm>WA</fnm></au>
    <au><snm>Asif</snm><fnm>A</fnm></au>
    <au><snm>Gull</snm><fnm>S</fnm></au>
    <au><snm>Minhas</snm><fnm>FUAA</fnm></au>
  </aug>
  <source>Journal of bioinformatics and computational biology</source>
  <publisher>World Scientific</publisher>
  <pubdate>2018</pubdate>
  <volume>16</volume>
  <issue>04</issue>
  <fpage>1850014</fpage>
</bibl>

<bibl id="B34">
  <title><p>Unified rational protein engineering with sequence-based deep
  representation learning</p></title>
  <aug>
    <au><snm>Alley</snm><fnm>EC</fnm></au>
    <au><snm>Khimulya</snm><fnm>G</fnm></au>
    <au><snm>Biswas</snm><fnm>S</fnm></au>
    <au><snm>AlQuraishi</snm><fnm>M</fnm></au>
    <au><snm>Church</snm><fnm>GM</fnm></au>
  </aug>
  <source>Nature methods</source>
  <publisher>Nature Publishing Group</publisher>
  <pubdate>2019</pubdate>
  <volume>16</volume>
  <issue>12</issue>
  <fpage>1315</fpage>
  <lpage>-1322</lpage>
</bibl>

<bibl id="B35">
  <title><p>Determining confidence of predicted interactions between HIV-1 and
  human proteins using conformal method</p></title>
  <aug>
    <au><snm>Nouretdinov</snm><fnm>I</fnm></au>
    <au><snm>Gammerman</snm><fnm>A</fnm></au>
    <au><snm>Qi</snm><fnm>Y</fnm></au>
    <au><snm>Klein Seetharaman</snm><fnm>J</fnm></au>
  </aug>
  <source>Biocomputing 2012</source>
  <publisher>World Scientific</publisher>
  <fpage>311</fpage>
  <lpage>-322</lpage>
</bibl>

<bibl id="B36">
  <title><p>Computational prediction of virus--human protein--protein
  interactions using embedding kernelized heterogeneous data</p></title>
  <aug>
    <au><snm>Nourani</snm><fnm>E</fnm></au>
    <au><snm>Khunjush</snm><fnm>F</fnm></au>
    <au><snm>Durmu{\c{s}}</snm><fnm>S</fnm></au>
  </aug>
  <source>Molecular BioSystems</source>
  <publisher>Royal Society of Chemistry</publisher>
  <pubdate>2016</pubdate>
  <volume>12</volume>
  <issue>6</issue>
  <fpage>1976</fpage>
  <lpage>-1986</lpage>
</bibl>

<bibl id="B37">
  <title><p>A novel one-class SVM based negative data sampling method for
  reconstructing proteome-wide HTLV-human protein interaction
  networks</p></title>
  <aug>
    <au><snm>Mei</snm><fnm>S</fnm></au>
    <au><snm>Zhu</snm><fnm>H</fnm></au>
  </aug>
  <source>Scientific reports</source>
  <publisher>Nature Publishing Group</publisher>
  <pubdate>2015</pubdate>
  <volume>5</volume>
  <issue>1</issue>
  <fpage>1</fpage>
  <lpage>-13</lpage>
</bibl>

<bibl id="B38">
  <title><p>Prediction of protein-protein interactions between viruses and
  human by an SVM model</p></title>
  <aug>
    <au><snm>Cui</snm><fnm>G</fnm></au>
    <au><snm>Fang</snm><fnm>C</fnm></au>
    <au><snm>Han</snm><fnm>K</fnm></au>
  </aug>
  <source>BMC bioinformatics</source>
  <pubdate>2012</pubdate>
  <volume>13</volume>
  <issue>7</issue>
  <fpage>1</fpage>
  <lpage>-10</lpage>
</bibl>

<bibl id="B39">
  <title><p>An improved method for predicting interactions between virus and
  human proteins</p></title>
  <aug>
    <au><snm>Kim</snm><fnm>B</fnm></au>
    <au><snm>Alguwaizani</snm><fnm>S</fnm></au>
    <au><snm>Zhou</snm><fnm>X</fnm></au>
    <au><snm>Huang</snm><fnm>DS</fnm></au>
    <au><snm>Park</snm><fnm>B</fnm></au>
    <au><snm>Han</snm><fnm>K</fnm></au>
  </aug>
  <source>Journal of bioinformatics and computational biology</source>
  <publisher>World Scientific</publisher>
  <pubdate>2017</pubdate>
  <volume>15</volume>
  <issue>01</issue>
  <fpage>1650024</fpage>
</bibl>

<bibl id="B40">
  <title><p>PredHPI: an integrated web server platform for the detection and
  visualization of host--pathogen interactions using sequence-based
  methods</p></title>
  <aug>
    <au><snm>Loaiza</snm><fnm>CD</fnm></au>
    <au><snm>Kaundal</snm><fnm>R</fnm></au>
  </aug>
  <source>Bioinformatics</source>
  <pubdate>2020</pubdate>
</bibl>

<bibl id="B41">
  <title><p>A generalized approach to predicting protein-protein interactions
  between virus and host</p></title>
  <aug>
    <au><snm>Zhou</snm><fnm>X</fnm></au>
    <au><snm>Park</snm><fnm>B</fnm></au>
    <au><snm>Choi</snm><fnm>D</fnm></au>
    <au><snm>Han</snm><fnm>K</fnm></au>
  </aug>
  <source>BMC genomics</source>
  <publisher>BioMed Central</publisher>
  <pubdate>2018</pubdate>
  <volume>19</volume>
  <issue>6</issue>
  <fpage>69</fpage>
  <lpage>-77</lpage>
</bibl>

<bibl id="B42">
  <title><p>Seq-BEL: Sequence-based Ensemble Learning for Predicting
  Virus-human Protein-protein Interaction</p></title>
  <aug>
    <au><snm>Ma</snm><fnm>Y</fnm></au>
    <au><snm>He</snm><fnm>T</fnm></au>
    <au><snm>Tan</snm><fnm>YT</fnm></au>
    <au><cnm>others</cnm></au>
  </aug>
  <source>IEEE/ACM Transactions on Computational Biology and
  Bioinformatics</source>
  <publisher>IEEE</publisher>
  <pubdate>2020</pubdate>
</bibl>

<bibl id="B43">
  <title><p>Predict the Protein-protein Interaction between Virus and Host
  through Hybrid Deep Neural Network</p></title>
  <aug>
    <au><snm>Deng</snm><fnm>L</fnm></au>
    <au><snm>Zhao</snm><fnm>J</fnm></au>
    <au><snm>Zhang</snm><fnm>J</fnm></au>
  </aug>
  <source>2020 IEEE International Conference on Bioinformatics and Biomedicine
  (BIBM)</source>
  <pubdate>2020</pubdate>
  <fpage>11</fpage>
  <lpage>-16</lpage>
</bibl>

<bibl id="B44">
  <title><p>Machine learning techniques for sequence-based prediction of
  viral--host interactions between SARS-CoV-2 and human proteins</p></title>
  <aug>
    <au><snm>Dey</snm><fnm>L</fnm></au>
    <au><snm>Chakraborty</snm><fnm>S</fnm></au>
    <au><snm>Mukhopadhyay</snm><fnm>A</fnm></au>
  </aug>
  <source>Biomedical journal</source>
  <publisher>Elsevier</publisher>
  <pubdate>2020</pubdate>
  <volume>43</volume>
  <issue>5</issue>
  <fpage>438</fpage>
  <lpage>-450</lpage>
</bibl>

<bibl id="B45">
  <title><p>Prediction of human-virus protein-protein interactions through a
  sequence embedding-based machine learning method</p></title>
  <aug>
    <au><snm>Yang</snm><fnm>X</fnm></au>
    <au><snm>Yang</snm><fnm>S</fnm></au>
    <au><snm>Li</snm><fnm>Q</fnm></au>
    <au><snm>Wuchty</snm><fnm>S</fnm></au>
    <au><snm>Zhang</snm><fnm>Z</fnm></au>
  </aug>
  <source>Computational and structural biotechnology journal</source>
  <publisher>Elsevier</publisher>
  <pubdate>2020</pubdate>
  <volume>18</volume>
  <fpage>153</fpage>
  <lpage>-161</lpage>
</bibl>

<bibl id="B46">
  <title><p>Transfer Learning for Predicting Virus-Host Protein Interactions
  for Novel Virus Sequences</p></title>
  <aug>
    <au><snm>Lanchantin</snm><fnm>J</fnm></au>
    <au><snm>Weingarten</snm><fnm>T</fnm></au>
    <au><snm>Sekhon</snm><fnm>A</fnm></au>
    <au><snm>Miller</snm><fnm>C</fnm></au>
    <au><snm>Qi</snm><fnm>Y</fnm></au>
  </aug>
  <source>bioRxiv</source>
  <publisher>Cold Spring Harbor Laboratory</publisher>
  <pubdate>2021</pubdate>
  <fpage>2020</fpage>
  <lpage>-12</lpage>
</bibl>

<bibl id="B47">
  <title><p>DeepViral: prediction of novel virus–host interactions from
  protein sequences and infectious disease phenotypes</p></title>
  <aug>
    <au><snm>Liu Wei</snm><fnm>W</fnm></au>
    <au><snm>Kafkas</snm><fnm>S</fnm></au>
    <au><snm>Chen</snm><fnm>J</fnm></au>
    <au><snm>Dimonaco</snm><fnm>NJ</fnm></au>
    <au><snm>Tegnér</snm><fnm>J</fnm></au>
    <au><snm>Hoehndorf</snm><fnm>R</fnm></au>
  </aug>
  <source>Bioinformatics</source>
  <pubdate>2021</pubdate>
</bibl>

<bibl id="B48">
  <title><p>Prediction of interactions between viral and host proteins using
  supervised machine learning methods</p></title>
  <aug>
    <au><snm>Barman</snm><fnm>RK</fnm></au>
    <au><snm>Saha</snm><fnm>S</fnm></au>
    <au><snm>Das</snm><fnm>S</fnm></au>
  </aug>
  <source>PloS one</source>
  <publisher>Public Library of Science</publisher>
  <pubdate>2014</pubdate>
  <volume>9</volume>
  <issue>11</issue>
  <fpage>e112034</fpage>
</bibl>

<bibl id="B49">
  <title><p>A structure-informed atlas of human-virus interactions</p></title>
  <aug>
    <au><snm>Lasso</snm><fnm>G</fnm></au>
    <au><snm>Mayer</snm><fnm>SV</fnm></au>
    <au><snm>Winkelmann</snm><fnm>ER</fnm></au>
    <au><snm>Chu</snm><fnm>T</fnm></au>
    <au><snm>Elliot</snm><fnm>O</fnm></au>
    <au><snm>Patino Galindo</snm><fnm>JA</fnm></au>
    <au><snm>Park</snm><fnm>K</fnm></au>
    <au><snm>Rabadan</snm><fnm>R</fnm></au>
    <au><snm>Honig</snm><fnm>B</fnm></au>
    <au><snm>Shapira</snm><fnm>SD</fnm></au>
  </aug>
  <source>Cell</source>
  <publisher>Elsevier</publisher>
  <pubdate>2019</pubdate>
  <volume>178</volume>
  <issue>6</issue>
  <fpage>1526</fpage>
  <lpage>-1541</lpage>
</bibl>

<bibl id="B50">
  <title><p>Predicting virus-host association by Kernelized logistic matrix
  factorization and similarity network fusion</p></title>
  <aug>
    <au><snm>Liu</snm><fnm>D</fnm></au>
    <au><snm>Ma</snm><fnm>Y</fnm></au>
    <au><snm>Jiang</snm><fnm>X</fnm></au>
    <au><snm>He</snm><fnm>T</fnm></au>
  </aug>
  <source>BMC bioinformatics</source>
  <publisher>Springer</publisher>
  <pubdate>2019</pubdate>
  <volume>20</volume>
  <issue>16</issue>
  <fpage>1</fpage>
  <lpage>-10</lpage>
</bibl>

<bibl id="B51">
  <title><p>A network-based integrated framework for predicting
  virus--prokaryote interactions</p></title>
  <aug>
    <au><snm>Wang</snm><fnm>W</fnm></au>
    <au><snm>Ren</snm><fnm>J</fnm></au>
    <au><snm>Tang</snm><fnm>K</fnm></au>
    <au><snm>Dart</snm><fnm>E</fnm></au>
    <au><snm>Ignacio Espinoza</snm><fnm>JC</fnm></au>
    <au><snm>Fuhrman</snm><fnm>JA</fnm></au>
    <au><snm>Braun</snm><fnm>J</fnm></au>
    <au><snm>Sun</snm><fnm>F</fnm></au>
    <au><snm>Ahlgren</snm><fnm>NA</fnm></au>
  </aug>
  <source>NAR genomics and bioinformatics</source>
  <publisher>Oxford University Press</publisher>
  <pubdate>2020</pubdate>
  <volume>2</volume>
  <issue>2</issue>
  <fpage>lqaa044</fpage>
</bibl>

<bibl id="B52">
  <title><p>Principles of Machine Learning-Guided Protein
  Engineering</p></title>
  <aug>
    <au><snm>Biswas</snm><fnm>S</fnm></au>
  </aug>
  <source>PhD thesis</source>
  <pubdate>2020</pubdate>
</bibl>

<bibl id="B53">
  <title><p>STRING v10: protein--protein interaction networks, integrated over
  the tree of life</p></title>
  <aug>
    <au><snm>Szklarczyk</snm><fnm>D</fnm></au>
    <au><snm>Franceschini</snm><fnm>A</fnm></au>
    <au><snm>Wyder</snm><fnm>S</fnm></au>
    <au><snm>Forslund</snm><fnm>K</fnm></au>
    <au><snm>Heller</snm><fnm>D</fnm></au>
    <au><snm>Huerta Cepas</snm><fnm>J</fnm></au>
    <au><snm>Simonovic</snm><fnm>M</fnm></au>
    <au><snm>Roth</snm><fnm>A</fnm></au>
    <au><snm>Santos</snm><fnm>A</fnm></au>
    <au><snm>Tsafou</snm><fnm>KP</fnm></au>
    <au><cnm>others</cnm></au>
  </aug>
  <source>Nucleic acids research</source>
  <publisher>Oxford University Press</publisher>
  <pubdate>2015</pubdate>
  <volume>43</volume>
  <issue>D1</issue>
  <fpage>D447</fpage>
  <lpage>-D452</lpage>
</bibl>

<bibl id="B54">
  <title><p>APID interactomes: providing proteome-based interactomes with
  controlled quality for multiple species and derived networks</p></title>
  <aug>
    <au><snm>Alonso Lopez</snm><fnm>D</fnm></au>
    <au><snm>Guti{\'e}rrez</snm><fnm>MA</fnm></au>
    <au><snm>Lopes</snm><fnm>KP</fnm></au>
    <au><snm>Prieto</snm><fnm>C</fnm></au>
    <au><snm>Santamar{\'\i}a</snm><fnm>R</fnm></au>
    <au><snm>De Las Rivas</snm><fnm>J</fnm></au>
  </aug>
  <source>Nucleic acids research</source>
  <publisher>Oxford University Press</publisher>
  <pubdate>2016</pubdate>
  <volume>44</volume>
  <issue>W1</issue>
  <fpage>W529</fpage>
  <lpage>-W535</lpage>
</bibl>

<bibl id="B55">
  <title><p>UniProt: a hub for protein information</p></title>
  <aug>
    <au><snm>Consortium</snm><fnm>U</fnm></au>
  </aug>
  <source>Nucleic acids research</source>
  <publisher>Oxford University Press</publisher>
  <pubdate>2015</pubdate>
  <volume>43</volume>
  <issue>D1</issue>
  <fpage>D204</fpage>
  <lpage>-D212</lpage>
</bibl>

<bibl id="B56">
  <title><p>PSICQUIC and PSISCORE: accessing and scoring molecular
  interactions</p></title>
  <aug>
    <au><snm>Aranda</snm><fnm>B</fnm></au>
    <au><snm>Blankenburg</snm><fnm>H</fnm></au>
    <au><snm>Kerrien</snm><fnm>S</fnm></au>
    <au><snm>Brinkman</snm><fnm>FS</fnm></au>
    <au><snm>Ceol</snm><fnm>A</fnm></au>
    <au><snm>Chautard</snm><fnm>E</fnm></au>
    <au><snm>Dana</snm><fnm>JM</fnm></au>
    <au><snm>De Las Rivas</snm><fnm>J</fnm></au>
    <au><snm>Dumousseau</snm><fnm>M</fnm></au>
    <au><snm>Galeota</snm><fnm>E</fnm></au>
    <au><cnm>others</cnm></au>
  </aug>
  <source>Nature methods</source>
  <publisher>Nature Publishing Group</publisher>
  <pubdate>2011</pubdate>
  <volume>8</volume>
  <issue>7</issue>
  <fpage>528</fpage>
  <lpage>-529</lpage>
</bibl>

<bibl id="B57">
  <title><p>Predicting protein--protein interactions using signature
  products</p></title>
  <aug>
    <au><snm>Martin</snm><fnm>S</fnm></au>
    <au><snm>Roe</snm><fnm>D</fnm></au>
    <au><snm>Faulon</snm><fnm>JL</fnm></au>
  </aug>
  <source>Bioinformatics</source>
  <publisher>Oxford University Press</publisher>
  <pubdate>2005</pubdate>
  <volume>21</volume>
  <issue>2</issue>
  <fpage>218</fpage>
  <lpage>-226</lpage>
</bibl>

<bibl id="B58">
  <title><p>Probability weighted ensemble transfer learning for predicting
  interactions between HIV-1 and human proteins</p></title>
  <aug>
    <au><snm>Mei</snm><fnm>S</fnm></au>
  </aug>
  <source>PLoS One</source>
  <publisher>Public Library of Science San Francisco, USA</publisher>
  <pubdate>2013</pubdate>
  <volume>8</volume>
  <issue>11</issue>
  <fpage>e79606</fpage>
</bibl>

<bibl id="B59">
  <title><p>The NCBI taxonomy database</p></title>
  <aug>
    <au><snm>Federhen</snm><fnm>S</fnm></au>
  </aug>
  <source>Nucleic acids research</source>
  <publisher>Oxford University Press</publisher>
  <pubdate>2012</pubdate>
  <volume>40</volume>
  <issue>D1</issue>
  <fpage>D136</fpage>
  <lpage>-D143</lpage>
</bibl>

<bibl id="B60">
  <title><p>Understanding eukaryotic linear motifs and their role in cell
  signaling and regulation</p></title>
  <aug>
    <au><snm>Diella</snm><fnm>F</fnm></au>
    <au><snm>Haslam</snm><fnm>N</fnm></au>
    <au><snm>Chica</snm><fnm>C</fnm></au>
    <au><snm>Budd</snm><fnm>A</fnm></au>
    <au><snm>Michael</snm><fnm>S</fnm></au>
    <au><snm>Brown</snm><fnm>NP</fnm></au>
    <au><snm>Trav{\'e}</snm><fnm>G</fnm></au>
    <au><snm>Gibson</snm><fnm>TJ</fnm></au>
  </aug>
  <source>Front Biosci</source>
  <publisher>Citeseer</publisher>
  <pubdate>2008</pubdate>
  <volume>13</volume>
  <issue>6580</issue>
  <fpage>603</fpage>
</bibl>

<bibl id="B61">
  <title><p>Peptides mediating interaction networks: new leads at
  last</p></title>
  <aug>
    <au><snm>Neduva</snm><fnm>V</fnm></au>
    <au><snm>Russell</snm><fnm>RB</fnm></au>
  </aug>
  <source>Current opinion in biotechnology</source>
  <publisher>Elsevier</publisher>
  <pubdate>2006</pubdate>
  <volume>17</volume>
  <issue>5</issue>
  <fpage>465</fpage>
  <lpage>-471</lpage>
</bibl>

<bibl id="B62">
  <title><p>Distributed representations of sentences and documents</p></title>
  <aug>
    <au><snm>Le</snm><fnm>Q</fnm></au>
    <au><snm>Mikolov</snm><fnm>T</fnm></au>
  </aug>
  <source>International conference on machine learning</source>
  <pubdate>2014</pubdate>
  <fpage>1188</fpage>
  <lpage>-1196</lpage>
</bibl>

<bibl id="B63">
  <title><p>Pytorch: An imperative style, high-performance deep learning
  library</p></title>
  <aug>
    <au><snm>Paszke</snm><fnm>A</fnm></au>
    <au><snm>Gross</snm><fnm>S</fnm></au>
    <au><snm>Massa</snm><fnm>F</fnm></au>
    <au><snm>Lerer</snm><fnm>A</fnm></au>
    <au><snm>Bradbury</snm><fnm>J</fnm></au>
    <au><snm>Chanan</snm><fnm>G</fnm></au>
    <au><snm>Killeen</snm><fnm>T</fnm></au>
    <au><snm>Lin</snm><fnm>Z</fnm></au>
    <au><snm>Gimelshein</snm><fnm>N</fnm></au>
    <au><snm>Antiga</snm><fnm>L</fnm></au>
    <au><cnm>others</cnm></au>
  </aug>
  <source>Advances in neural information processing systems</source>
  <pubdate>2019</pubdate>
  <volume>32</volume>
  <fpage>8026</fpage>
  <lpage>-8037</lpage>
</bibl>

<bibl id="B64">
  <title><p>The generalization of ‘STUDENT'S’problem when several different
  population varlances are involved</p></title>
  <aug>
    <au><snm>Welch</snm><fnm>BL</fnm></au>
  </aug>
  <source>Biometrika</source>
  <publisher>Oxford University Press</publisher>
  <pubdate>1947</pubdate>
  <volume>34</volume>
  <issue>1-2</issue>
  <fpage>28</fpage>
  <lpage>-35</lpage>
</bibl>

<bibl id="B65">
  <title><p>On comparing classifiers: Pitfalls to avoid and a recommended
  approach</p></title>
  <aug>
    <au><snm>Salzberg</snm><fnm>SL</fnm></au>
  </aug>
  <source>Data mining and knowledge discovery</source>
  <publisher>Springer</publisher>
  <pubdate>1997</pubdate>
  <volume>1</volume>
  <issue>3</issue>
  <fpage>317</fpage>
  <lpage>-328</lpage>
</bibl>

<bibl id="B66">
  <title><p>Handbook of parametric and nonparametric statistical
  procedures</p></title>
  <aug>
    <au><snm>Kafadar</snm><fnm>K</fnm></au>
  </aug>
  <source>The American Statistician</source>
  <publisher>American Statistical Association</publisher>
  <pubdate>1997</pubdate>
  <volume>51</volume>
  <issue>4</issue>
  <fpage>374</fpage>
</bibl>

<bibl id="B67">
  <title><p>A mass spectrometric-derived cell surface protein atlas</p></title>
  <aug>
    <au><snm>Bausch Fluck</snm><fnm>D</fnm></au>
    <au><snm>Hofmann</snm><fnm>A</fnm></au>
    <au><snm>Bock</snm><fnm>T</fnm></au>
    <au><snm>Frei</snm><fnm>AP</fnm></au>
    <au><snm>Cerciello</snm><fnm>F</fnm></au>
    <au><snm>Jacobs</snm><fnm>A</fnm></au>
    <au><snm>Moest</snm><fnm>H</fnm></au>
    <au><snm>Omasits</snm><fnm>U</fnm></au>
    <au><snm>Gundry</snm><fnm>RL</fnm></au>
    <au><snm>Yoon</snm><fnm>C</fnm></au>
    <au><cnm>others</cnm></au>
  </aug>
  <source>PloS one</source>
  <publisher>Public Library of Science</publisher>
  <pubdate>2015</pubdate>
  <volume>10</volume>
  <issue>4</issue>
  <fpage>e0121314</fpage>
</bibl>

<bibl id="B68">
  <title><p>Gene ontology: tool for the unification of biology</p></title>
  <aug>
    <au><snm>Ashburner</snm><fnm>M</fnm></au>
    <au><snm>Ball</snm><fnm>CA</fnm></au>
    <au><snm>Blake</snm><fnm>JA</fnm></au>
    <au><snm>Botstein</snm><fnm>D</fnm></au>
    <au><snm>Butler</snm><fnm>H</fnm></au>
    <au><snm>Cherry</snm><fnm>JM</fnm></au>
    <au><snm>Davis</snm><fnm>AP</fnm></au>
    <au><snm>Dolinski</snm><fnm>K</fnm></au>
    <au><snm>Dwight</snm><fnm>SS</fnm></au>
    <au><snm>Eppig</snm><fnm>JT</fnm></au>
    <au><cnm>others</cnm></au>
  </aug>
  <source>Nature genetics</source>
  <publisher>Nature Publishing Group</publisher>
  <pubdate>2000</pubdate>
  <volume>25</volume>
  <issue>1</issue>
  <fpage>25</fpage>
  <lpage>-29</lpage>
</bibl>

<bibl id="B69">
  <title><p>The Gene Ontology resource: enriching a GOld mine</p></title>
  <aug>
    <au><snm>Carbon</snm><fnm>S</fnm></au>
    <au><snm>Douglass</snm><fnm>E</fnm></au>
    <au><snm>Good</snm><fnm>BM</fnm></au>
    <au><snm>Unni</snm><fnm>DR</fnm></au>
    <au><snm>Harris</snm><fnm>NL</fnm></au>
    <au><snm>Mungall</snm><fnm>CJ</fnm></au>
    <au><snm>Basu</snm><fnm>S</fnm></au>
    <au><snm>Chisholm</snm><fnm>RL</fnm></au>
    <au><snm>Dodson</snm><fnm>RJ</fnm></au>
    <au><snm>Hartline</snm><fnm>E</fnm></au>
    <au><cnm>others</cnm></au>
  </aug>
  <source>Nucleic Acids Research</source>
  <publisher>Oxford University Press (OUP)</publisher>
  <pubdate>2021</pubdate>
  <volume>49</volume>
  <issue>D1</issue>
  <fpage>D325</fpage>
  <lpage>-D334</lpage>
</bibl>

<bibl id="B70">
  <title><p>Cell entry mechanisms of SARS-CoV-2</p></title>
  <aug>
    <au><snm>Shang</snm><fnm>J</fnm></au>
    <au><snm>Wan</snm><fnm>Y</fnm></au>
    <au><snm>Luo</snm><fnm>C</fnm></au>
    <au><snm>Ye</snm><fnm>G</fnm></au>
    <au><snm>Geng</snm><fnm>Q</fnm></au>
    <au><snm>Auerbach</snm><fnm>A</fnm></au>
    <au><snm>Li</snm><fnm>F</fnm></au>
  </aug>
  <source>Proceedings of the National Academy of Sciences</source>
  <publisher>National Acad Sciences</publisher>
  <pubdate>2020</pubdate>
  <volume>117</volume>
  <issue>21</issue>
  <fpage>11727</fpage>
  <lpage>-11734</lpage>
</bibl>

<bibl id="B71">
  <title><p>Molecular mechanism of interaction between SARS-CoV-2 and host
  cells and interventional therapy</p></title>
  <aug>
    <au><snm>Zhang</snm><fnm>Q</fnm></au>
    <au><snm>Xiang</snm><fnm>R</fnm></au>
    <au><snm>Huo</snm><fnm>S</fnm></au>
    <au><snm>Zhou</snm><fnm>Y</fnm></au>
    <au><snm>Jiang</snm><fnm>S</fnm></au>
    <au><snm>Wang</snm><fnm>Q</fnm></au>
    <au><snm>Yu</snm><fnm>F</fnm></au>
  </aug>
  <source>Signal Transduction and Targeted Therapy</source>
  <publisher>Nature Publishing Group</publisher>
  <pubdate>2021</pubdate>
  <volume>6</volume>
  <issue>1</issue>
  <fpage>1</fpage>
  <lpage>-19</lpage>
</bibl>

<bibl id="B72">
  <title><p>SARS-CoV-2 cell entry depends on ACE2 and TMPRSS2 and is blocked by
  a clinically proven protease inhibitor</p></title>
  <aug>
    <au><snm>Hoffmann</snm><fnm>M</fnm></au>
    <au><snm>Kleine Weber</snm><fnm>H</fnm></au>
    <au><snm>Schroeder</snm><fnm>S</fnm></au>
    <au><snm>Kr{\"u}ger</snm><fnm>N</fnm></au>
    <au><snm>Herrler</snm><fnm>T</fnm></au>
    <au><snm>Erichsen</snm><fnm>S</fnm></au>
    <au><snm>Schiergens</snm><fnm>TS</fnm></au>
    <au><snm>Herrler</snm><fnm>G</fnm></au>
    <au><snm>Wu</snm><fnm>NH</fnm></au>
    <au><snm>Nitsche</snm><fnm>A</fnm></au>
    <au><cnm>others</cnm></au>
  </aug>
  <source>cell</source>
  <publisher>Elsevier</publisher>
  <pubdate>2020</pubdate>
  <volume>181</volume>
  <issue>2</issue>
  <fpage>271</fpage>
  <lpage>-280</lpage>
</bibl>

<bibl id="B73">
  <title><p>Angiotensin-converting enzyme 2 is a functional receptor for the
  SARS coronavirus</p></title>
  <aug>
    <au><snm>Li</snm><fnm>W</fnm></au>
    <au><snm>Moore</snm><fnm>MJ</fnm></au>
    <au><snm>Vasilieva</snm><fnm>N</fnm></au>
    <au><snm>Sui</snm><fnm>J</fnm></au>
    <au><snm>Wong</snm><fnm>SK</fnm></au>
    <au><snm>Berne</snm><fnm>MA</fnm></au>
    <au><snm>Somasundaran</snm><fnm>M</fnm></au>
    <au><snm>Sullivan</snm><fnm>JL</fnm></au>
    <au><snm>Luzuriaga</snm><fnm>K</fnm></au>
    <au><snm>Greenough</snm><fnm>TC</fnm></au>
    <au><cnm>others</cnm></au>
  </aug>
  <source>Nature</source>
  <publisher>Nature Publishing Group</publisher>
  <pubdate>2003</pubdate>
  <volume>426</volume>
  <issue>6965</issue>
  <fpage>450</fpage>
  <lpage>-454</lpage>
</bibl>

<bibl id="B74">
  <title><p>The pathogenicity of SARS-CoV-2 in hACE2 transgenic
  mice</p></title>
  <aug>
    <au><snm>Bao</snm><fnm>L</fnm></au>
    <au><snm>Deng</snm><fnm>W</fnm></au>
    <au><snm>Huang</snm><fnm>B</fnm></au>
    <au><snm>Gao</snm><fnm>H</fnm></au>
    <au><snm>Liu</snm><fnm>J</fnm></au>
    <au><snm>Ren</snm><fnm>L</fnm></au>
    <au><snm>Wei</snm><fnm>Q</fnm></au>
    <au><snm>Yu</snm><fnm>P</fnm></au>
    <au><snm>Xu</snm><fnm>Y</fnm></au>
    <au><snm>Qi</snm><fnm>F</fnm></au>
    <au><cnm>others</cnm></au>
  </aug>
  <source>Nature</source>
  <publisher>Nature Publishing Group</publisher>
  <pubdate>2020</pubdate>
  <volume>583</volume>
  <issue>7818</issue>
  <fpage>830</fpage>
  <lpage>-833</lpage>
</bibl>

<bibl id="B75">
  <title><p>SARS-CoV-2 infection of human ACE2-transgenic mice causes severe
  lung inflammation and impaired function</p></title>
  <aug>
    <au><snm>Winkler</snm><fnm>ES</fnm></au>
    <au><snm>Bailey</snm><fnm>AL</fnm></au>
    <au><snm>Kafai</snm><fnm>NM</fnm></au>
    <au><snm>Nair</snm><fnm>S</fnm></au>
    <au><snm>McCune</snm><fnm>BT</fnm></au>
    <au><snm>Yu</snm><fnm>J</fnm></au>
    <au><snm>Fox</snm><fnm>JM</fnm></au>
    <au><snm>Chen</snm><fnm>RE</fnm></au>
    <au><snm>Earnest</snm><fnm>JT</fnm></au>
    <au><snm>Keeler</snm><fnm>SP</fnm></au>
    <au><cnm>others</cnm></au>
  </aug>
  <source>Nature immunology</source>
  <publisher>Nature Publishing Group</publisher>
  <pubdate>2020</pubdate>
  <volume>21</volume>
  <issue>11</issue>
  <fpage>1327</fpage>
  <lpage>-1335</lpage>
</bibl>

<bibl id="B76">
  <title><p>Structure of MERS-CoV spike receptor-binding domain complexed with
  human receptor DPP4</p></title>
  <aug>
    <au><snm>Wang</snm><fnm>N</fnm></au>
    <au><snm>Shi</snm><fnm>X</fnm></au>
    <au><snm>Jiang</snm><fnm>L</fnm></au>
    <au><snm>Zhang</snm><fnm>S</fnm></au>
    <au><snm>Wang</snm><fnm>D</fnm></au>
    <au><snm>Tong</snm><fnm>P</fnm></au>
    <au><snm>Guo</snm><fnm>D</fnm></au>
    <au><snm>Fu</snm><fnm>L</fnm></au>
    <au><snm>Cui</snm><fnm>Y</fnm></au>
    <au><snm>Liu</snm><fnm>X</fnm></au>
    <au><cnm>others</cnm></au>
  </aug>
  <source>Cell research</source>
  <publisher>Nature Publishing Group</publisher>
  <pubdate>2013</pubdate>
  <volume>23</volume>
  <issue>8</issue>
  <fpage>986</fpage>
  <lpage>-993</lpage>
</bibl>

<bibl id="B77">
  <title><p>Emerging COVID-19 coronavirus: glycan shield and structure
  prediction of spike glycoprotein and its interaction with human
  CD26</p></title>
  <aug>
    <au><snm>Vankadari</snm><fnm>N</fnm></au>
    <au><snm>Wilce</snm><fnm>JA</fnm></au>
  </aug>
  <source>Emerging microbes \& infections</source>
  <publisher>Taylor \& Francis</publisher>
  <pubdate>2020</pubdate>
  <volume>9</volume>
  <issue>1</issue>
  <fpage>601</fpage>
  <lpage>-604</lpage>
</bibl>

<bibl id="B78">
  <title><p>Human aminopeptidase N is a receptor for human coronavirus
  229E</p></title>
  <aug>
    <au><snm>Yeager</snm><fnm>CL</fnm></au>
    <au><snm>Ashmun</snm><fnm>RA</fnm></au>
    <au><snm>Williams</snm><fnm>RK</fnm></au>
    <au><snm>Cardellichio</snm><fnm>CB</fnm></au>
    <au><snm>Shapiro</snm><fnm>LH</fnm></au>
    <au><snm>Look</snm><fnm>AT</fnm></au>
    <au><snm>Holmes</snm><fnm>KV</fnm></au>
  </aug>
  <source>Nature</source>
  <publisher>Nature Publishing Group</publisher>
  <pubdate>1992</pubdate>
  <volume>357</volume>
  <issue>6377</issue>
  <fpage>420</fpage>
  <lpage>-422</lpage>
</bibl>

<bibl id="B79">
  <title><p>Highly accurate protein structure prediction with
  AlphaFold</p></title>
  <aug>
    <au><snm>Jumper</snm><fnm>J</fnm></au>
    <au><snm>Evans</snm><fnm>R</fnm></au>
    <au><snm>Pritzel</snm><fnm>A</fnm></au>
    <au><snm>Green</snm><fnm>T</fnm></au>
    <au><snm>Figurnov</snm><fnm>M</fnm></au>
    <au><snm>Ronneberger</snm><fnm>O</fnm></au>
    <au><snm>Tunyasuvunakool</snm><fnm>K</fnm></au>
    <au><snm>Bates</snm><fnm>R</fnm></au>
    <au><snm>{\v{Z}}{\'\i}dek</snm><fnm>A</fnm></au>
    <au><snm>Potapenko</snm><fnm>A</fnm></au>
    <au><cnm>others</cnm></au>
  </aug>
  <source>Nature</source>
  <publisher>Nature Publishing Group</publisher>
  <pubdate>2021</pubdate>
  <fpage>1</fpage>
  <lpage>-11</lpage>
</bibl>

<bibl id="B80">
  <title><p>A multitask transfer learning framework for novel virus-human
  protein interactions</p></title>
  <aug>
    <au><snm>Dong</snm><fnm>NT</fnm></au>
    <au><snm>Khosla</snm><fnm>M</fnm></au>
  </aug>
  <source>bioRxiv</source>
  <publisher>Cold Spring Harbor Laboratory</publisher>
  <pubdate>2021</pubdate>
</bibl>

</refgrp>
} 

\clearpage
\end{backmatter}
\appendix 
\section{Appendix}

\subsection{Detailed Results}
The following subsections provide detailed experimental results. For the \textsc{Hybrid} and \motif, the author's code is not available and results are taken from the original paper as the. `-' indicates that the score is not available. For other methods, the reported results are the average after 10 experimental runs. We perform pairwise t-test tests for statistical significance testing. Our presented results are statistically significant with a p-value less than 0.05.

\subsection{Comparison with Methods using hand crafted protein features}
Table \ref{tab:hardcode} provides a comparison between \mtt and baselines which employ hand-crafted features. \mtt outperfroms \denovo in all benchmarked datasets  while \mtt supersede \generalized in six out of the seven datasets. The performance gains are statistically significant with a p-value of 0.05.  
\begin{table}[h!]
    \centering
    \caption{Comparison with methods based on hand-crafted features}
    \begin{tabular}{llccccc}
         \textsc{Dataset}& \textsc{Model} & \textsc{AUC} & \textsc{AP} &  \textsc{Precision} &  \textsc{Recall} & \textsc{F1} \\
        \midrule
        
         \multirow{3}{*}{\zhouh} & \textsc{Denovo}  & $0.8656$ & $0.8619$ & $77.75$ & $77.95$ & $77.85$ \\
         & \generalized  & $0.8600$ & $0.8606$ & $76.96$ & $77.17$ & $77.06$ \\
           & \textsc{Hybrid} 
         & $0.937$ & - & - & - & - \\
         & \mtt& $\textbf{0.9461}$ & $\textbf{0.9589}$ & $\textbf{86.28}$ & $\textbf{86.51}$ & $\textbf{86.40}$ \\
         \midrule
         \multirow{3}{*}{\zhoue} & \textsc{Denovo}  & $0.8864$ & $0.8366$ & $83.44$ & $84.00$ & $83.72$\\
         & \textsc{Generalized}  & $0.9154$ & $0.9078$ & $84.77$ & $85.33$ & $85.05$ \\
         & \mtt& $\textbf{0.9680}$ & $\textbf{0.9766}$ & $\textbf{90.93}$ & $\textbf{91.53}$ & $\textbf{91.23}$ \\
         \midrule
         \multirow{3}{*}{\slim} & \textsc{Denovo}  & $0.8701$ & $0.8631$ & $81.92$ & $82.12$ & $82.02$ \\
        & \generalized  & $0.8891$ & $0.8851$ & $\textbf{84.74}$ & $\textbf{84.94}$ & $\textbf{84.84}$ \\
         &  \mtt& $\textbf{0.9221}$ & $\textbf{0.9324}$ & $83.92$ & $84.12$ & $84.02$ \\
         \midrule
         \multirow{3}{*}{\barman} & \textsc{Denovo}  & $0.8217$ & $0.8415$ & $74.60$ & $74.98$ & $74.79$ \\
         & \generalized  & $0.8214$ & $0.8458$ & $74.90$ & $75.27$ & $75.08$ \\
         &\mtt& $\textbf{0.9804}$ & $\textbf{0.9802}$ & $\textbf{93.53}$ & $\textbf{94.05}$ & $\textbf{93.79}$ \\
         \midrule
         \multirow{3}{*}{\textsc{Bacillus}} & \textsc{Denovo}  & $0.9843$ & $0.9650$ & $94.80$ & $94.83$ & $94.83$ \\
         &\generalized  & $0.9833$ & $0.9668$ & $95.75$ & $95.78$ & $95.76$ \\
         & \mtt& $\textbf{0.9997}$ & $\textbf{0.9992}$ & $\textbf{98.75}$ & $\textbf{98.78}$ & $\textbf{98.76}$\\
         \midrule
         \multirow{3}{*}{\textsc{Yersina}} & \denovo
         & $0.9712$ & $0.9302$ & $93.14$ & $93.16$ & $93.15$  \\
         &\generalized  & $0.9758$ & $0.9362$ & $94.01$ & $94.03$ & $94.02$ \\
         & \mtt& $\textbf{0.9988}$ & $\textbf{0.9971}$ & $\textbf{97.32}$ & $\textbf{97.34}$ & $\textbf{97.32}$ \\
         \midrule
         \multirow{3}{*}{\textsc{Franci}} &  \textsc{Denovo}  & $0.9782$ & $0.9584$ & $95.55$ & $95.62$ & $95.58$ \\
         &\generalized  & $0.9799$ & $0.9565$ & $95.84$ & $95.91$ & $95.88$ \\
         & \mtt& $\textbf{0.9998}$ & $\textbf{0.9996}$ & $\textbf{98.95}$ & $\textbf{99.03}$ & $\textbf{98.99}$ \\
        \end{tabular}
    
    \label{tab:hardcode}
    \end{table}
\subsection{Comparison with sequence embedding based Methods}
Table \ref{tab:embedding} provides  a comparison between \mtt and embedding-based methods on small testing datasets. \mtt outperforms \docvec in 4 datasets. The performance gains are statistically significant with a p-value of 0.05. \mtt is outperformed by \docvec in three datasets: \zhoue, \zhouh and \slim. We point out that these datasets are quite specialized where the negative training and testing samples were drawn from a sequence dissimilarity negative sampling technique. In particular, the protein sequences for negative test set were already chosen based on their dissimilarity to those is positive set. This is not a realistic setting when the positive test set is in itself unknown. Nevertheless, the performance of \mtt is comparable to the state of the art on these especially curated datasets too while it outperforms all methods on more general datasets.
\begin{table}[h!]
    \centering
    \caption{Comparison with embedding-based methods}
    \begin{tabular}{llccccc}
         \textsc{Dataset}& \textsc{Model} & \textsc{AUC} & \textsc{AP} &  \textsc{Precision} &  \textsc{Recall} & \textsc{F1} \\
        \midrule
        \multirow{3}{*}{\zhouh} & \textsc{doc2vec} & $\textbf{0.9601}$ & $\textbf{0.9674}$ & $\textbf{89.04}$ & $\textbf{89.34}$ & $\textbf{89.19}$ \\
         & \motif  & $0.945$ & $-$ & - & - & $86.50$ \\
         & \mtt& $0.9461$ & $0.9589$ & $86.28$ & $86.51$ & $86.40$ \\
         \midrule
         \multirow{3}{*}{\zhoue}&\textsc{Doc2vec}  & $\textbf{0.9781}$ & $\textbf{0.9832}$ & $\textbf{91.99}$ & $\textbf{92.67}$ & $\textbf{92.33}$ \\
         & \motif & $0.968$ & $-$ & - & - & $89.6$ \\
      & \mtt& $0.9680$ & $0.9766$ & $90.93$ & $91.53$ & $91.23$ \\ 
         \midrule
         \multirow{2}{*}{\slim} &    \textsc{doc2vec} & $\textbf{0.9644}$ & $\textbf{0.9681}$ & $\textbf{88.60}$ & $\textbf{88.87}$ & $\textbf{88.73}$ \\
         &  \mtt& $0.9221$ & $0.9324$ & $83.92$ & $84.12$ & $84.02$ \\
         \midrule
         \multirow{2}{*}{\barman} & \textsc{doc2vec} & $0.8671$ & $0.8922$ & $79.95$ & $80.37$ & $80.16$ \\
         &\mtt& $\textbf{0.9804}$ & $\textbf{0.9802}$ & $\textbf{93.53}$ & $\textbf{94.05}$ & $\textbf{93.79}$ \\
         \midrule
         \multirow{2}{*}{\textsc{Bacillus}} & \textsc{doc2vec} & $0.9900$ & $0.9739$ & $96.29$ & $96.32$ & $96.31$ \\
         & \mtt& $\textbf{0.9997}$ & $\textbf{0.9992}$ & $\textbf{98.75}$ & $\textbf{98.78}$ & $\textbf{98.76}$ \\
         \midrule
         \multirow{2}{*}{\textsc{Yersina}}& \textsc{doc2vec} & $0.9814$ & $0.9510$ & $94.50$ & $94.52$ & $94.51$ \\
         & \mtt& $\textbf{0.9988}$ & $\textbf{0.9971}$ & $\textbf{97.32}$ & $\textbf{97.34}$ & $\textbf{97.32}$ \\
         \midrule
         \multirow{2}{*}{\textsc{Franci}}  & \textsc{doc2vec} & $0.9878$ & $0.9606$ & $96.77$ & $96.84$ & $96.81$ \\
         & \mtt& $\textbf{0.9998}$ & $\textbf{0.9996}$ & $\textbf{98.95}$ & $\textbf{99.03}$ & $\textbf{98.99}$ \\
        \end{tabular}
    
    \label{tab:embedding}
    \end{table}

\subsection{Detailed Results on Ablation Studies}
\begin{table}[h!]
\centering
\caption{Results for Ablation Studies.}
 \label{tab:ablation}
    \begin{tabular}{llcccccc}
         \textsc{Dataset} & \textsc{Model} & \textsc{AUC} & \textsc{AP} &  \textsc{Precision} &  \textsc{Recall} & \textsc{F1} \\
        \toprule
         \multirow{3}{*}{\textsc{H1N1} }
        & \naive& $0.8310$ & $0.8003$ & $75.92$ & $76.12$ & $76.02$ \\ 
        & \stt& $0.9472$ & $0.9590$ & $85.86$ & $86.09$ & $85.98$ \\
        & \mtt& $\textbf{0.9461}$ & $\textbf{0.9589}$ & $\textbf{86.28}$ & $\textbf{86.51}$ & $\textbf{86.40}$ \\
      
         \midrule 
         \multirow{3}{*}{\textsc {Ebola}} 
     & \naive& $0.8876$ & $0.8665$ & $82.12$ & $82.67$ & $82.39$ \\ 
      & \stt& $0.9655$ & $0.9749$ &  $90.13$ & $90.73$ & $90.43$ \\
      & \mtt& $\textbf{0.9680}$ & $\textbf{0.9766}$ & $90.93$ & $91.53$ & $91.23$ \\ 
     
      \midrule
      \multirow{3}{*}{\slim}
   
     & \naive& $0.8843$ & $0.8673$ & $83.80$ & $84.00$ & $83.90$ \\ 
      &  \stt& $0.9207$ & $0.9343$ &  $84.04$ & $84.24$ & $84.14$ \\
      &  \mtt& $\textbf{0.9221}$ & $\textbf{0.9324}$ & $\textbf{83.92}$ & $\textbf{84.12}$ & $\textbf{84.02}$ \\
    
      \midrule
        \multirow{3}{*}{\barman's}&\naive& $0.8084$ & $0.8198$ & $73.75$ & $74.11$ & $73.93$ \\ 
      &\mtt& $0.9804$ & $0.9802$ & $93.53$ & $94.05$ & $93.79$ \\
      &\stt& $0.9801$ & $0.9802$ & $93.83$ & $94.29$ & $94.06$ \\
    
      \midrule
        \multirow{3}{*}{\textsc{Bacillus}} 
     & \naive& $0.9842$ & $0.9619$ & $93.75$ & $93.78$ & $93.77$ \\ 
      & \stt& $0.9995$ & $0.9986$ & $97.93$ & $97.96$ & $97.95$ \\
      & \mtt& $\textbf{0.9997}$ & $\textbf{0.9992}$ & $\textbf{98.75}$ & $\textbf{98.78}$ & $\textbf{98.76}$ \\
   
      \midrule
      \multirow{3}{*}{\textsc{Yersina}} 
        
         & \naive& $0.9741$ & $0.9277$ & $92.61$ & $92.64$ & $92.63$ \\ 
      & \stt& $\textbf{0.9987}$ & $0.9970$ & $97.18$ & $97.30$ & $97.24$ \\
       & \mtt& $0.9988$ & $\textbf{0.9971}$ & $97.32$ & $97.34$ & $97.32$ \\
     
      \midrule
      \multirow{3}{*}{\textsc{Franci}   } 
        & \naive& $0.9851$ & $0.9680$ & $94.36$ & $94.43$ & $94.39$ \\ 
      & \stt& $0.9997$ & $0.9993$ & $98.84$ & $98.92$ & $98.88$ \\
      & \mtt& $\textbf{0.9998}$ & $\textbf{0.9996}$ & $98.95$ & $\textbf{99.03}$ & $\textbf{98.99}$ \\
  
    \midrule
    \multirow{3}{*}{\textsc{2697049}} 
    & \naive& $0.5686$ & $0.0010$ & $0$ & $0$ & $0$ \\ 
      & \stt& $0.7457$ & $0.0017$ & $0.07$ & $0.07$ & $0.07$  \\
    & \mtt& $\textbf{0.7566}$ & $\textbf{0.0021}$ & $\textbf{0.97}$ & $\textbf{0.97}$ & $\textbf{0.97}$\\
    
      \midrule
      \multirow{3}{*}{\textsc{333761}}
    
      & \naive& $0.7002$ & $0.0110$ & $3.55$ & $3.56$ & $3.55$ \\ 
      & \stt& $0.8114$ & $0.0213$ & $4.72$ & $4.72$ & $4.72$  \\ 
      & \mtt& $0.8160$ & $0.0262$ & $\textbf{6.35}$ & $6.35$ & $\textbf{6.35}$ \\ 
    
      \midrule
      \multirow{3}{*}{\textsc{2043570}} 
    
    & \naive& $0.6624$ & $0.0076$ & $0.32$ & $0.32$ & $0.32$ \\ 
      & \stt& $0.6706$ & $0.0087$ & $1.11$ & $\textbf{3.01}$ & $1.46$\\ 
      & \mtt& $\textbf{0.6956}$ & $\textbf{0.0096}$ & $\textbf{1.89}$ & $1.91$ & $\textbf{1.90}$ \\ 
    
      \midrule
      \multirow{3}{*}{\textsc{644788}}
    
      & \naive& $0.8410$ & $0.0089$ & $1.82$ & $1.85$ & $1.83$\\ 
      & \stt& $\textbf{0.9705}$ & $\textbf{0.0459}$ & $3.97$ & $9.26$ & $4.65$\\ 
      & \mtt& $0.9537$ & $0.0302$  & $3.54$ & $\textbf{22.04}$ & $5.46$ \\ 
    \end{tabular}
      
\end{table}
In Table \ref{tab:ablation} we compare \mtt with its simpler variants for 11 datasets. The reported results are average after 10 runs. Results from pair-wise t-test show that that (i)\mtt is significantly better than \naive in all datasets with p-value smaller than 0.05, (ii) \mtt is significantly better than \stt in 4 out of 7 datasets with p-value smaller than 0.05 while for the remaining datasets, the difference is not statistically significant.

\end{document}